\documentclass[10pt,twocolumn,letterpaper]{article}
\usepackage[pagenumbers]{cvpr}

\usepackage{cuted} 

\usepackage{currfile} 

\usepackage{caption} 

\usepackage{microtype}

\usepackage{booktabs}
\usepackage{placeins}
\usepackage{multirow}
\usepackage{amsmath,amssymb}
\usepackage{xcolor}
\usepackage{graphicx}
\usepackage{tikz}
\usetikzlibrary{positioning,calc,arrows.meta}
\definecolor{cvprblue}{rgb}{0.21,0.49,0.74}
\usepackage[breaklinks,hidelinks]{hyperref}

\newcommand{\method}{\textup{\textsc{Homeward}}}
\newcommand{\pizero}{$\pi_0$}
\newcommand{\pifive}{$\pi_{0.5}$}

\def\confName{CVPR}
\def\confYear{2027}

\title{\textsc{Homeward}: No Place Like the Training View}
\author{\begin{minipage}[t]{0.96\textwidth}\normalfont\centering\fontsize{10.5}{13}\selectfont
\mbox{Mingyang Lyu$^{1,2,3}$}\quad \mbox{Yinqian Sun$^{1}$}\quad \mbox{Gongce Cao$^{3}$}\quad \mbox{Guobin Shen$^{6}$}\\[2pt]\mbox{Yuhao Zhang$^{4}$}\quad \mbox{Lu Jia$^{5}$}\quad \mbox{Chenfei Yan$^{7}$}\quad \mbox{Qian Zhang$^{1}$}\\[2pt]\mbox{Pin Lyu$^{1,\dagger}$}\quad \mbox{Chen Gao$^{8,\dagger}$}\quad \mbox{Bo Wang$^{1,\dagger}$}
\\[5pt]{\fontsize{8.5}{10.5}\selectfont 
$^{1}$Institute of Automation, Chinese Academy of Sciences, Beijing, China\\$^{2}$Zhongguancun Academy, Beijing, China\\$^{3}$University of Chinese Academy of Sciences, Beijing, China\\$^{4}$University of Illinois Urbana-Champaign, USA\\$^{5}$University of Science and Technology Beijing, Beijing, China\\$^{6}$ByteDance, China\\$^{7}$Beihang University, Beijing, China\\$^{8}$Tsinghua University, Beijing, China
\\[4pt]$^\dagger$Corresponding authors: Pin Lyu, Chen Gao, Bo Wang.}
\end{minipage}
}

\hypersetup{pdftitle={Homeward: No Place Like the Training View},pdfauthor={Mingyang Lyu, Yinqian Sun, Gongce Cao, Guobin Shen, Yuhao Zhang, Lu Jia, Chenfei Yan, Qian Zhang, Pin Lyu, Chen Gao, Bo Wang}}
\begin{document}
\raggedbottom
\twocolumn[{%
\renewcommand\twocolumn[1][]{#1}%
\maketitle
\vspace{-1.2em}
\begin{center}
\captionsetup{hypcap=false}
\includegraphics[width=0.9\linewidth]{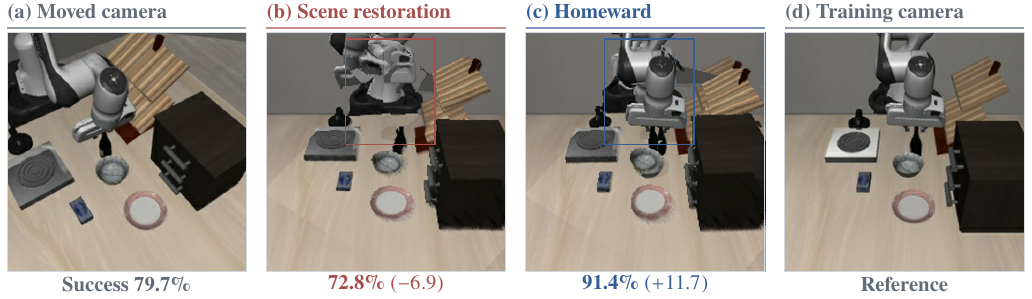}
\captionof{figure}{\textbf{Naive 3D scene restoration can leave a frozen VLA worse off than no correction---unless paired with kinematic robot rendering.} \textbf{(a)} Displacing the third-person camera ($34^\circ$ azimuth, $15^\circ$ elevation) degrades frozen OpenVLA-OFT. \textbf{(b)} Reprojecting the observation with a frozen 3D foundation model tears and scatters the moving arm (red box), \emph{lowering} success from 79.7\% to 72.8\%. \textbf{(c)} \method{} composites moving links rendered from measured joint states (blue box) over the restored scene, boosting success to \textbf{91.4\%}. \textbf{(d)} The canonical training-camera view, for reference. Percentages report closed-loop OpenVLA-OFT success across all 408 LIBERO-Goal camera variants.}
\label{fig:teaser}
\end{center}
\vspace{0.5em}
}]
\begin{abstract}
Vision-language-action (VLA) policies implicitly assume that third-person cameras remain fixed at their data-collection poses; once a camera is bumped, remounted, or swapped at deployment, task success degrades precipitously even though the underlying task and robot remain unchanged.
We present \method{}, a training-free, plug-and-play observation adapter that canonicalizes observations from an uncalibrated deployment camera back into the training view for frozen VLA policies.
Our design stems from a closed-loop diagnostic paradox in simulation: while re-rendering perturbed views into the training camera with oracle geometry recovers much of the lost performance, substituting an off-the-shelf feed-forward 3D foundation model can leave the policy \emph{worse off} than applying no correction at all.
This degradation concentrates on the thin, articulated robot arm, which feed-forward depth prediction tears and scatters---yet the arm's exact 3D state is already known from robot kinematics.
\method{} exploits this asymmetry: it \emph{estimates what is unknown}---camera parameters and static scene depth---from the live RGB frame and a single training-view reference, and \emph{renders what is known}---the moving robot links---directly from the kinematic model.
Evaluated across $\sim$300,000 closed-loop episodes spanning four frozen VLAs, \method{} boosts success across all 1,599 camera variants of LIBERO-Plus from 62.5\%, 76.5\%, 68.0\%, and 3.6\% to 77.6\%, 95.1\%, 91.2\%, and 26.4\% for OpenVLA-OFT, \pifive{}, \pizero{}, and OpenVLA, respectively.
Without any scene fine-tuning or deployment camera calibration, \method{} extends to 6-DoF random placements, focal-length changes, and, with per-step pose estimation, continuous intra-episode camera drift, while matching or surpassing a calibrated, scene-trained view-synthesis baseline on three of four policies on LIBERO-Plus.
\end{abstract}

\section{Introduction}
\label{sec:intro}

Vision-language-action (VLA) models have rapidly advanced generalist robotic manipulation by mapping raw visual observations and natural language instructions directly to actions~\cite{brohan2023rt1,zitkovich2023rt2,octo2024,oneill2024openx,kim2024openvla,kim2025oft,black2024pi0,intelligence2025pi05}.
Yet beneath their broad semantic generalization lies a brittle geometric assumption inherited from data collection: at deployment, the third-person camera must remain precisely where it stood during demonstrations.
In practice, this assumption breaks routinely when cameras are bumped, re-mounted, or replaced with different lenses.
Although neither the task nor the robot has changed, the unseen viewpoint severely disrupts action prediction~\cite{xie2023decomposing}.
Indeed, LIBERO-Plus~\cite{fei2025liberoplus} identifies camera viewpoint shifts as the single most damaging perturbation for frozen VLAs; across its 1,599 camera variants, OpenVLA-OFT~\cite{kim2025oft} drops to 62.5\% overall success and plummets to 22.0\% under the largest shifts (\cref{fig:levels}).

One natural remedy is to bake viewpoint invariance into the policy itself via multi-view data augmentation~\cite{chen2024roviaug,tian2024vista}, cross-view consistency objectives~\cite{huang2026crossview,li2026camvla}, 3D-aware architectures~\cite{abouzeid2025geoaware,wilcox2025adapt3r}, or few-shot adaptation in the target view~\cite{li2025ftm}.
However, retraining or modifying large pre-trained VLAs sacrifices modularity and must be repeated for every policy family.
An appealing orthogonal paradigm is \emph{test-time observation canonicalization}: leaving the policy completely untouched and transforming incoming images back into the training viewpoint.
Yet existing test-time adapters merely shift the engineering burden to deployment---requiring extrinsic calibration of the moved camera, metric depth sensors, or generative view-synthesis networks fine-tuned on target scenes~\cite{gu2026vistabot,heo2026anycamvla,pan2026corevla,park2026gsvla,sathua2026argus}.
Because only the observer's coordinate frame has changed, we ask: \emph{How far can a frozen VLA be recovered from arbitrary camera shifts when the deployment camera is uncalibrated and nothing is trained?}

Closed-loop simulation reveals a striking paradox.
Reprojecting perturbed observations into the training view using privileged simulator depth and ground-truth camera poses restores OpenVLA-OFT on LIBERO-Goal from 79.7\% to 90.4\% overall---and from 35.5\% to 82.9\% under the severest shifts---confirming that viewpoint failure resides predominantly in visual input space.
Yet when we replace privileged geometry with an off-the-shelf 3D foundation model (VGGT~\cite{wang2025vggt}), closed-loop success drops to 72.8\%---\emph{worse} than feeding the unaligned view directly to the policy (\cref{fig:teaser}).
Diagnostic inspection pinpoints the culprit: while feed-forward 3D models recover static background layout reasonably well, they tear and scatter the thin, specular, moving robot arm.
Prior work noticed this reconstruction bottleneck and responded by fine-tuning 3D backbones on robot datasets~\cite{gu2026vistabot}.
Our key insight is simpler: \emph{the robot arm does not need to be visually reconstructed at all}, because its exact 3D geometry is already given at every timestep by the kinematic model and joint proprioception.

Building on this asymmetric decomposition between unknown scene geometry and known robot kinematics, we introduce \method{}, a training-free test-time observation adapter.
Given a single clean reference frame from the training camera, \method{} leverages a frozen 3D foundation model to estimate the deployment camera parameters and current scene depth, reprojects the workspace into the training view, inpaints disocclusions from the reference frame, and composites the articulated robot links rendered directly from measured joint states.
While kinematic robot cross-painting has previously enabled cross-embodiment transfer under \emph{calibrated} setups~\cite{chen2024mirage,lepert2024shadow}, \method{} demonstrates that pairing kinematic rendering with feed-forward monocular geometry unlocks reliable view restoration with \textbf{a frozen policy, a frozen geometry backbone, an uncalibrated deployment camera, and zero adaptation data}---requiring only the robot model, one reference image, and a one-time training-camera calibration.
Restoring this kinematic layer elevates OpenVLA-OFT on LIBERO-Goal from 72.8\% to 91.4\%, matching privileged oracle re-rendering (90.4\%).

Evaluated over roughly 300,000 closed-loop rollouts, our contributions are threefold:
\begin{itemize}
\item \textbf{Closed-Loop Viewpoint Diagnostic.} We quantify how much frozen VLA failure under camera shifts is recoverable purely in observation space, and demonstrate that naive restoration with off-the-shelf feed-forward 3D models can underperform uncorrected inputs due to foreground reconstruction artifacts on the moving arm (\cref{sec:diagnosis}).
\item \textbf{Training-Free Kinematic--Geometric Canonicalization.} We propose \method{}, which decouples observation restoration into monocular scene reprojection and forward-kinematic link rendering, with initial-pose caching and motion gating for low-overhead deployment without deployment-camera calibration (\cref{sec:scene,sec:robot,sec:caching}).
\item \textbf{Large-Scale Evaluation \& Regime Characterization.} Across four frozen VLAs and five camera perturbation families---including 1,599 LIBERO-Plus variants ($+15.1$ to $+23.2$ points), 6-DoF random poses, continuous drift, and focal-length changes---we show where zero-shot geometric restoration excels, how it complements fine-tuned policies out-of-distribution, and where wide-baseline calibrated view synthesis remains ahead (\cref{sec:exp}).
\end{itemize}

\section{Related Work}
\label{sec:related}
\begin{figure*}[t]
\centering
\includegraphics[width=0.8\linewidth]{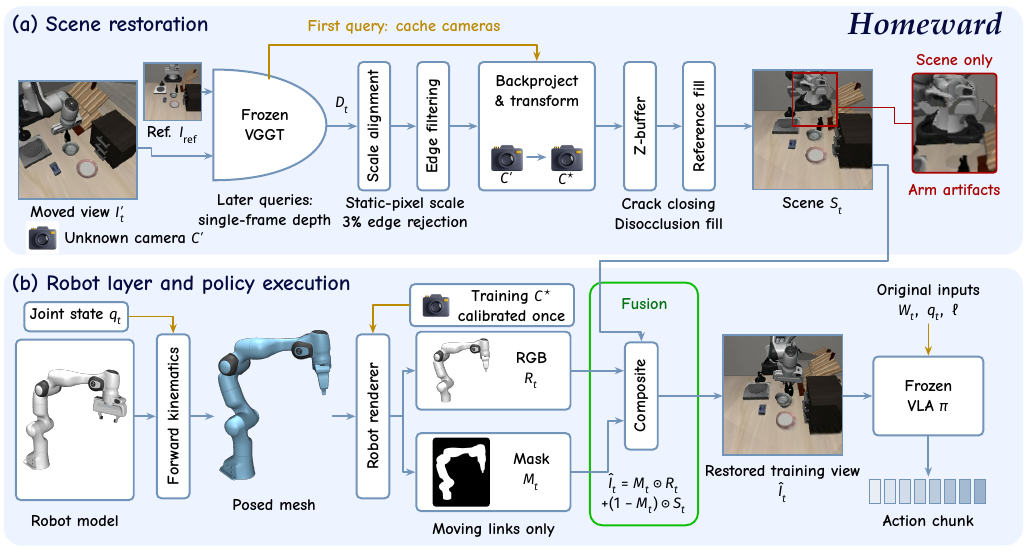}
\caption{\textbf{Overview of \method{}.} \textbf{(a) Monocular scene restoration:} At the first policy query, frozen VGGT estimates and caches the deployment camera transform ($C' \to C^\star$) from $[I_{\mathrm{ref}}, I'_0]$; subsequent queries infer single-frame depth $D_t$ aligned on static pixels. Depth-edge filtering, z-buffer splatting, and disocclusion filling from $I_{\mathrm{ref}}$ yield the scene layer $S_t$, which retains artifacts on the moving arm (inset). \textbf{(b) Kinematic robot layer \& policy execution:} Forward kinematics and the one-time training camera calibration $C^\star$ render moving-link RGB $R_t$ and silhouette mask $M_t$, composited via $\hat I_t = M_t \odot R_t + (1 - M_t) \odot S_t$ and passed with the unchanged $W_t$, $q_t$, and $\ell$ to the frozen VLA $\pi$.}
\label{fig:method}
\end{figure*}

\paragraph{Training-Time Viewpoint Robustness for Robot Policies.}
Generalization benchmarks probe camera-pose perturbations~\cite{xie2023decomposing,pumacay2024colosseum}, and LIBERO-Plus~\cite{fei2025liberoplus} identifies third-person viewpoint shifts as a primary failure mode of modern VLAs.
Most existing solutions intervene during policy training: synthesizing novel viewpoints for data augmentation~\cite{chen2024roviaug,tian2024vista}, fine-tuning on multi-view perturbed trajectories~\cite{fei2025liberoplus}, lifting observations into calibrated 3D/RGB-D representations~\cite{shridhar2022peract,goyal2023rvt,goyal2024rvt2,ke2024diffuseractor,ze2024dp3,abouzeid2025geoaware,wilcox2025adapt3r}, designing view-invariant architectures~\cite{li2026camvla}, or enforcing cross-view action consistency~\cite{huang2026crossview}.
Alternatively, FTM~\cite{li2025ftm} adapts visual encoders at test time using a demonstration collected in the new camera view.
While effective within their target distributions, all such methods either modify policy weights or require new-view demonstrations, whereas \method{} operates strictly in input space for off-the-shelf frozen policies.

\paragraph{Test-Time Observation Canonicalization.}
A complementary line of work transforms test-time observations back to the training camera frame (\cref{tab:req}).
\emph{Neural view-synthesis adapters} rely on scene- or robot-specific training: AnyCamVLA~\cite{heo2026anycamvla} fine-tunes a large view-synthesis model~\cite{jin2024lvsm} on LIBERO-Plus renders conditioned on known third-person and wrist camera parameters, while GS-VLA~\cite{park2026gsvla} canonicalizes views via a trained Gaussian-splatting~\cite{kerbl2023gaussian} module given deployment poses and depth.
VistaBot~\cite{gu2026vistabot} pairs VGGT~\cite{wang2025vggt} with a video diffusion synthesizer; notably, it also observes that feed-forward 3D models struggle on robot arms and grippers, addressing this by fine-tuning both VGGT and downstream policy features on robot data.
\emph{Geometric reprojection and cross-painting approaches} avoid generative hallucination by leveraging known sensor calibration: CoRe-VLA~\cite{pan2026corevla} reprojects calibrated RGB-D frames followed by a learned restoration network, ARGUS~\cite{sathua2026argus} fuses two calibrated views for policies trained on canonicalized inputs, and Mirage~\cite{chen2024mirage} masks target robots and renders source-robot meshes from joint states across calibrated cameras for cross-embodiment transfer.
\method{} unifies monocular estimated geometry with kinematic robot rendering under a strictly weaker deployment assumption: \textbf{zero model training or fine-tuning, and an uncalibrated monocular deployment camera}.
We identify the design choices that make this combination viable in closed loop (\cref{sec:mechanism}).

\begin{table}[t]
\centering\footnotesize\setlength{\tabcolsep}{3pt}
\begin{tabular}{@{}lccc@{}}
\toprule
& trained or & moved & robot \\
& fine-tuned & camera & region \\
\midrule
AnyCamVLA~\cite{heo2026anycamvla} & synthesis model & calibrated & synthesized \\
VistaBot~\cite{gu2026vistabot} & VGGT, policy & intrinsics & reconstructed \\
Mirage~\cite{chen2024mirage} & dynamics model & calibrated & rendered \\
\textbf{\method{}} & \textbf{none} & \textbf{uncalibrated} & \textbf{rendered} \\
\bottomrule
\end{tabular}

\caption{\textbf{Deployment assumptions of test-time viewpoint adapters.} \emph{Trained or fine-tuned} lists modules trained beyond public weights; \emph{Moved camera} indicates calibration required at deployment. All but VistaBot assume one-time calibration of the training camera $C^\star$ (Mirage targets cross-embodiment transfer).}
\label{tab:req}
\end{table}

\paragraph{Feed-Forward 3D Foundation Models.}
Beyond single-image depth and geometry estimators~\cite{yang2024depthanythingv2,hu2024metric3dv2,wang2025moge}, multi-view transformers such as DUSt3R~\cite{wang2024dust3r}, MASt3R~\cite{leroy2024mast3r}, VGGT~\cite{wang2025vggt}, and $\pi^3$~\cite{wang2025pi3} infer camera intrinsics, extrinsics, and dense aligned depth from uncalibrated RGB images in a single forward pass without bundle adjustment.
In manipulation, however, a specular arm that moves between the reference and live frames violates the static-scene assumption of two-view geometry~\cite{hartley2004multiple,zhang2025monst3r}.
Rather than fine-tuning 3D foundation models on robot data, we characterize how their failure modes impact closed-loop VLA control and bypass foreground errors via forward kinematics.

\section{Method}
\label{sec:method}

\subsection{Problem Formulation}
We consider a frozen VLA policy $\pi$ that maps a multimodal observation $o_t=(I_t,W_t,q_t)$ and a language instruction $\ell$ to actions, where $I_t$ is the third-person RGB image, $W_t$ is the wrist-camera image, and $q_t$ denotes the robot joint configuration.
During deployment, the third-person camera is displaced from its training configuration $C^\star$ to an unknown configuration $C'$, yielding a perturbed image $I'_t$.
We assume access to three standard quantities available prior to deployment: (i) a single clean reference image $I_{\mathrm{ref}}$ of the workspace captured from $C^\star$, (ii) the robot's kinematic mesh model (\eg, a URDF), and (iii) the extrinsic calibration of the original training camera $C^\star$ relative to the robot base, recorded once during data collection.
Without calibrating $C'$ or modifying $\pi$, our objective is to reconstruct an estimate $\hat I_t$ of the canonical training view and feed $(\hat I_t,W_t,q_t,\ell)$ directly into the frozen policy.

\subsection{Closed-Loop Viewpoint Diagnostic}
\label{sec:diagnosis}
We first conduct a controlled closed-loop study on the LIBERO-Plus camera variants of LIBERO-Goal and LIBERO-Long to isolate what input-view canonicalization can---and cannot---recover (\cref{tab:prelim}; \cref{sec:setup}).

\begin{table}[t]
\centering\scriptsize\setlength{\tabcolsep}{2.4pt}
\begin{tabular}{@{}lcccccccc@{}}
\toprule
& \multicolumn{2}{c}{OFT} & \multicolumn{2}{c}{\pifive{}} & \multicolumn{2}{c}{\pizero{}} & \multicolumn{2}{c}{OpenVLA} \\
\cmidrule(lr){2-3}\cmidrule(lr){4-5}\cmidrule(lr){6-7}\cmidrule(l){8-9}
& G & L & G & L & G & L & G & L \\
\midrule
frozen policy & 79.7 & 35.3 & 84.3 & 52.7 & 65.2 & 44.9 & 8.1 & 1.2 \\
oracle re-rendering$^\dagger$ & 90.4$^\ast$ & 76.1$^\ast$ & 88.7 & 77.8$^\ast$ & 89.7$^\ast$ & 74.7$^\ast$ & 27.5$^\ast$ & 15.0$^\ast$ \\
oracle homography$^\dagger$ & 70.8$^\ast$ & 38.7 & 74.8$^\ast$ & 51.3 & 60.8 & 41.5 & 13.7$^\ast$ & 4.5$^\ast$ \\
\midrule
scene restoration only & 72.8$^\ast$ & 51.3$^\ast$ & 76.0$^\ast$ & 60.4$^\ast$ & 82.6$^\ast$ & 62.5$^\ast$ & 22.3$^\ast$ & 7.2$^\ast$ \\
\textbf{\method{}} & \textbf{91.4}$^\ast$ & \textbf{72.3}$^\ast$ & \textbf{94.4}$^\ast$ & \textbf{94.0}$^\ast$ & \textbf{93.4}$^\ast$ & \textbf{78.3}$^\ast$ & \textbf{39.2}$^\ast$ & \textbf{11.2}$^\ast$ \\
\bottomrule
\end{tabular}

\caption{\textbf{Closed-loop diagnostic study} (success rate, \%) on LIBERO-Plus camera variants of Goal (G, 408) and Long (L, 419). $^\dagger$Privileged, non-deployable (simulator depth and/or cameras). $^\ast$Exact McNemar $p<0.05$ \vs frozen policy.}
\label{tab:prelim}
\end{table}

\paragraph{Recoverable via 3D geometry, not 2D warping.}
Re-rendering every displaced observation $I'_t$ into $C^\star$ using privileged simulator depth and ground-truth camera poses (\emph{oracle re-rendering}) boosts closed-loop success on LIBERO-Goal by $+10.7$, $+4.4$, and $+24.5$ percentage points for OpenVLA-OFT, \pifive{}, and \pizero{}, respectively, and by $+25.1$ to $+40.8$ points on LIBERO-Long.
Action-level inspection reveals why: under a shifted camera, the policy's initial planned motion still points in roughly the right direction ($1$--$4^\circ$ angular error), but misjudges step magnitude by 17--28\%; oracle 3D re-rendering compresses this magnitude error to 3--10\%.
Conversely, warping $I'_t$ via a ground-truth plane-induced homography~\cite{hartley2004multiple} (\emph{oracle homography}) degrades success on Goal below the uncorrected policy for three of four models, since a single plane leaves off-plane parallax unresolved.

\paragraph{Off-the-shelf 3D estimation fails on the moving arm.}
Replacing privileged geometry with estimates from a frozen VGGT~\cite{wang2025vggt} given $[I_{\mathrm{ref}},I'_t]$ (\emph{scene restoration only}) causes OpenVLA-OFT (72.8\% \vs 79.7\%) and \pifive{} (76.0\% \vs 84.3\%) on Goal to drop \emph{below} the uncorrected frozen policy, while leaving \pizero{} far short of the oracle on both suites.
Visual inspection exposes a sharp dichotomy (\cref{fig:teaser}b): VGGT reprojects the static tabletop and objects plausibly, whereas the thin, moving robot arm is severely torn and scattered across depth planes.
Crucially, the articulated arm is the exact component whose 3D geometry is fully determined by $q_t$ and the kinematic model.
\method{} therefore decomposes $\hat I_t$ into two complementary layers (\cref{fig:method}): a \emph{scene layer} $S_t$ canonicalized via estimated monocular geometry, and a \emph{robot layer} $(R_t,M_t)$ rendered from exact forward kinematics.

\subsection{Monocular Scene Restoration}
\label{sec:scene}
Given the uncalibrated image pair $[I_{\mathrm{ref}},I'_t]$, frozen VGGT predicts camera intrinsics $(K_{\mathrm{ref}},K_t)$, extrinsics $(E_{\mathrm{ref}},E_t)$, and a dense depth map $D_t$ for $I'_t$ within a shared metric-up-to-scale coordinate frame.
We back-project each homogeneous pixel $\tilde u$ of $I'_t$ into 3D space and reproject it into the canonical training view $C^\star$:
\begin{equation}
u^\star = \Pi\!\left(K_{\mathrm{ref}}\, E_{\mathrm{ref}}\, E_t^{-1}\!\left[D_t(u)\, K_t^{-1}\tilde u\right]\right),
\end{equation}
where $\Pi(\cdot)$ is perspective projection and rigid transformations act on homogeneous coordinates.
Reprojected colors are forward-splatted at policy resolution with a z-buffer.

To suppress boundary ``flying pixels'' where feed-forward depth blends foreground and background across occlusion boundaries, we reject pixels whose depth deviates from any immediate neighbor by more than 3\% prior to splatting.
Isolated single-pixel rasterization cracks are interpolated from local neighbors, and remaining disoccluded regions are inpainted from $I_{\mathrm{ref}}$.
Across all LIBERO-Plus camera variants, live reprojected pixels cover 70\% of the canonical view on average---ranging from 88\% under mild estimated rotations ($<10^\circ$) to 59\% under wide ones ($>30^\circ$)---with $I_{\mathrm{ref}}$ supplying static background context for the remainder to form the restored scene layer $S_t$.

\subsection{Kinematic Robot Rendering and Compositing}
\label{sec:robot}
Because $S_t$ retains severe depth artifacts along thin, specular, moving manipulator links, we overwrite the foreground robot using forward kinematics.
Given joint state $q_t$ and the one-time training-camera calibration $C^\star$, a lightweight mesh renderer generates canonical robot RGB appearance $R_t$ and binary silhouette mask $M_t$, which are composited over $S_t$:
\begin{equation}
\hat I_t = M_t \odot R_t + (1-M_t) \odot S_t.
\end{equation}

\paragraph{Selective moving-link rendering.}
$(R_t,M_t)$ includes \emph{only} the articulated moving links, leaving the static robot base and mounting pedestal to the scene layer $S_t$.
Because the mesh renderer operates without full-scene occlusion geometry, rendering the entire robot---including the lower pedestal normally occluded by the tabletop in $C^\star$---paints over workspace pixels and collapses OpenVLA-OFT success on Goal to 39.0\% (\cref{tab:ablation}).
Furthermore, this layer tolerates imperfect calibration and appearance: perturbing $C^\star$ by $(10\,\mathrm{mm},1^\circ)$ costs at most 3.2 percentage points and $(20\,\mathrm{mm},2^\circ)$ leaves every evaluated cell above its uncorrected baseline, while drawing the arm with the color error measured for a real-robot render ($\approx$30 gray levels) costs at most 8.8 points on Goal and Long, with every cell still significantly above the frozen policy (supplement).

\subsection{Pose Caching, Motion Gating, and Runtime}
\label{sec:caching}
\paragraph{Pose caching.}
By default, we estimate and cache $(K_t,E_t)$ at the first policy query ($t=0$), which keeps correspondence matching away from later arm motion.
Subsequent queries run single-frame depth inference on $I'_t$ alone and rescale the prediction $\bar D_t$ to the cached initial depth $D_0$:
\begin{equation}
D_t = s_t\,\bar D_t,\qquad s_t = \operatorname*{median}_{u\in\Omega_t}\, D_0(u)/\bar D_t(u),
\label{eq:scale}
\end{equation}
where $\Omega_t$ collects the static pixels whose mean absolute color difference from $I'_0$ is below 12 of 255 intensity levels---decoupling depth scaling from arm articulation.
For scenarios involving continuous intra-episode camera motion, a \emph{per-step} variant re-estimates two-view pose at every query.

\paragraph{Do-no-harm motion gating.}
A motion gate checks the initial estimate at $t=0$: if estimated rotation is below $0.5^\circ$, translation is under 1\% of median scene depth, and focal length differs by less than 10\%, \method{} bypasses restoration and forwards raw frames untouched for the rest of the episode (the per-step variant re-tests at every query).

\paragraph{Runtime overhead.}
Using VGGT-1B (518\,px, bfloat16) on a single NVIDIA A100 GPU, cached scene restoration takes 0.16\,s per policy query and kinematic robot rendering requires 0.01\,s (per-step pose estimation adds 0.04\,s).
Estimated from the wall-clock difference between restored and frozen closed-loop rollouts, total overhead is $\sim$0.21\,s per query---amortizing to \textbf{27\,ms per control step} for the 8-step action-chunking policies~\cite{zhao2023act,chi2023diffusion}.
The simulator pauses during inference; letting the robot instead hold its pose for 5 control steps (0.25\,s) at every query changes success by $-3.7$ to $+1.2$ points on Goal and Long (supplement).

\section{Experiments}
\label{sec:exp}

\subsection{Experimental Setup}
\label{sec:setup}

\paragraph{Benchmarks and Camera Perturbations.}
We evaluate five camera-shift regimes alongside unperturbed controls.
(1)~\emph{LIBERO-Plus Camera}~\cite{fei2025liberoplus} perturbs the third-person camera across the four LIBERO suites~\cite{liu2023libero} in azimuth, elevation, distance, and in-place roll across five severity levels; we evaluate all 1,599 variants (one episode per variant from initial state 0).
(2)~\emph{AnyCamVLA Orbits}~\cite{heo2026anycamvla} orbit the camera around the workspace across small, medium, and large offsets (up to 15\,cm radial distance and $60^\circ$ azimuth), evaluated across 50 initial states per task on Long and 10 on the remaining suites.
(3)~\emph{CamRand} samples random 3D camera positions within spheres of three radii around $C^\star$, re-aiming at a jittered workspace target with random roll (300 episodes per suite; 1,200 total).
(4)~\emph{Camera Drift} continuously translates and rotates the camera from $C^\star$ to an AnyCamVLA orbital pose over the first 120 control steps (1,200 episodes).
(5)~\emph{Lens Change} alters the field of view (FoV) from $45^\circ$ to $55^\circ$ or $35^\circ$ at fixed extrinsics (800 episodes).
(6)~\emph{Unmoved Camera Controls} comprise 100 clean-camera episodes per suite and all 2,400 non-camera perturbation variants of LIBERO-Plus.

\paragraph{Evaluated Policies and Statistical Protocol.}
We test official LIBERO checkpoints of OpenVLA-OFT~\cite{kim2025oft}, \pifive{}~\cite{intelligence2025pi05}, \pizero{}~\cite{black2024pi0}, and OpenVLA~\cite{kim2024openvla} with strictly frozen weights, rendered via the GPU EGL backend of MuJoCo~\cite{todorov2012mujoco} (reproducing~\cite{fei2025liberoplus} under its official prompt convention; see supplement).
LIBERO-Plus, CamRand, and clean episodes use standard task instructions, while AnyCamVLA-orbit, drift, and lens-change benchmarks use LIBERO-Plus paraphrased instructions identically across all methods.
Each task's $I_{\mathrm{ref}}$ is rendered once from an initial state other than the evaluated one, so its object and arm poses differ from the test episode.
All comparisons are paired on identical episodes using exact McNemar tests~\cite{mcnemar1947}, supplemented by 95\% task-level cluster bootstrap~\cite{efron1979bootstrap} confidence intervals (reported in brackets) that resample tasks (40 on LIBERO-Plus) to account for shared scene geometry.
Because the gate never triggered in 4,797 displaced-camera and 600 lens-change trials, we report ungated cached execution and ungated per-step execution under drift. For OpenVLA-OFT, \pifive{}, and \pizero{}, per-query gating changes drift success by $-1.8$ to $+0.5$ points and stays within 0.8 points of frozen policies on clean cameras (supplement).

\begin{table*}[t]
\centering\small\setlength{\tabcolsep}{5pt}
\begin{tabular}{lrcccccccc}
\toprule
& & \multicolumn{2}{c}{OpenVLA-OFT} & \multicolumn{2}{c}{\pifive{}} & \multicolumn{2}{c}{\pizero{}} & \multicolumn{2}{c}{OpenVLA} \\
\cmidrule(lr){3-4}\cmidrule(lr){5-6}\cmidrule(lr){7-8}\cmidrule(lr){9-10}
Suite & Variants & frozen & \textbf{+\,\method{}} & frozen & \textbf{+\,\method{}} & frozen & \textbf{+\,\method{}} & frozen & \textbf{+\,\method{}} \\
\midrule
Spatial & 376 & 91.8 & \textbf{97.3}$^\ast$ & 76.9 & \textbf{98.1}$^\ast$ & 79.5 & \textbf{98.7}$^\ast$ & 3.2 & \textbf{32.2}$^\ast$ \\
Object & 396 & 45.7 & \textbf{50.3} & 93.2 & \textbf{93.9} & 84.6 & \textbf{95.7}$^\ast$ & 2.0 & \textbf{23.7}$^\ast$ \\
Goal & 408 & 79.7 & \textbf{91.4}$^\ast$ & 84.3 & \textbf{94.4}$^\ast$ & 65.2 & \textbf{93.4}$^\ast$ & 8.1 & \textbf{39.2}$^\ast$ \\
Long & 419 & 35.3 & \textbf{72.3}$^\ast$ & 52.7 & \textbf{94.0}$^\ast$ & 44.9 & \textbf{78.3}$^\ast$ & 1.2 & \textbf{11.2}$^\ast$ \\
\midrule
All & 1,599 & 62.5 & \textbf{77.6}$^\ast$ & 76.5 & \textbf{95.1}$^\ast$ & 68.0 & \textbf{91.2}$^\ast$ & 3.6 & \textbf{26.4}$^\ast$ \\
Gain & & \multicolumn{2}{c}{\textbf{+15.1}} & \multicolumn{2}{c}{\textbf{+18.6}} & \multicolumn{2}{c}{\textbf{+23.2}} & \multicolumn{2}{c}{\textbf{+22.8}} \\
\bottomrule
\end{tabular}

\caption{\textbf{LIBERO-Plus camera dimension.} Success rate (\%) of frozen policies on all 1,599 camera variants, one episode per variant, without and with \method{}. $^\ast$\,exact McNemar $p<0.05$ on the paired episodes; for ``All'', $p<10^{-31}$ for every policy. No policy is retrained and no calibration of the moved camera is given.}
\label{tab:main}
\end{table*}

\begin{figure*}[t]
\centering
\includegraphics[width=0.78\linewidth]{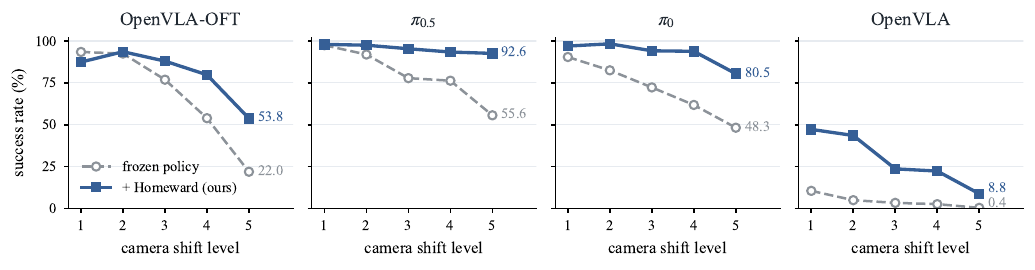}
\caption{\textbf{Closed-loop success across camera-shift severity levels}, pooled across all four LIBERO-Plus suites (levels 1--5 comprise 199, 360, 325, 228, and 487 variants). While frozen policies degrade steeply with viewpoint shift, \method{} degrades far less up to Level 5---lifting OpenVLA-OFT from 22.0\% to 53.8\%, \pifive{} from 55.6\% to 92.6\%, and \pizero{} from 48.3\% to 80.5\%. OpenVLA benefits most at Level 1 (10.6\% to 47.2\%).}
\label{fig:levels}
\end{figure*}

\subsection{Main Results on LIBERO-Plus}
\label{sec:main}
\paragraph{Consistent recovery across architectures and severities.}
Across all 1,599 camera variants of LIBERO-Plus (\cref{tab:main}), \method{} improves aggregate task success across every VLA---boosting OpenVLA-OFT, \pifive{}, \pizero{}, and OpenVLA by $+15.1$, $+18.6$, $+23.2$, and $+22.8$ points ($p<10^{-31}$ each).
Out of 16 suite--policy combinations, 14 improve significantly and none degrades significantly; the largest gains emerge on long-horizon manipulation (LIBERO-Long, $+33.4$ to $+41.3$ points across the first three policies).
Resampling the 40 tasks yields 95\% confidence intervals of $[7.6, 22.9]$, $[10.3, 27.1]$, $[17.1, 29.7]$, and $[17.7, 28.4]$ points, with per-task success improving on 26, 23, 35, and 38 of 40 tasks (and decreasing on only 5, 9, 1, and 1; task-level Wilcoxon signed-rank $p<0.01$ for every policy).

\paragraph{Graceful degradation under extreme camera shifts.}
\label{sec:generic}%
Stratifying results across perturbation severity levels 1--5 (\cref{fig:levels}) shows that while frozen OpenVLA-OFT, \pifive{}, and \pizero{} suffer catastrophic drops of 42--72 points from Level 1 to Level 5, \method{} restricts losses to 5--34 points.
At Level 1, where perturbations are minimal, restoration costs OpenVLA-OFT 6.1 points (87.4\% \vs 93.5\%).
Non-3D alternatives fall short: temporal ensembling and 2D registration to $I_{\mathrm{ref}}$ move OpenVLA-OFT, \pifive{}, and \pizero{} by at most 2.7 points. With the same robot layer, 2D registration and oracle homography trail \method{} by 3.4--26.5 points on Goal and Long (all 12 paired comparisons against \method{} significant; supplement).

\begin{table}[t]
\centering\small\setlength{\tabcolsep}{3.2pt}
\begin{tabular}{lcccc}
\toprule
& OFT & \pifive{} & \pizero{} & OpenVLA \\
\midrule
\multicolumn{5}{l}{\emph{CamRand}} \\
frozen & 83.8 & 93.9 & 88.2 & 30.5 \\
+\,\method{} & \textbf{93.6}$^\ast$ & \textbf{97.2}$^\ast$ & \textbf{93.0}$^\ast$ & \textbf{62.8}$^\ast$ \\
\midrule
\multicolumn{5}{l}{\emph{AnyCamVLA-style, other suites}} \\
frozen & 62.0 & 59.3 & 43.3 & 5.9 \\
+\,\method{} & \textbf{70.9}$^\ast$ & \textbf{79.7}$^\ast$ & \textbf{62.3}$^\ast$ & \textbf{12.6}$^\ast$ \\
\midrule
\multicolumn{5}{l}{\emph{Camera drift}} \\
frozen & 58.1 & 67.5 & 50.3 & 11.4 \\
+\,\method{} & \textbf{65.0}$^\ast$ & 66.1 & \textbf{55.0}$^\ast$ & 8.5$^\ast$ \\
+\,\method{}, per-step pose & \textbf{72.5}$^\ast$ & \textbf{82.5}$^\ast$ & \textbf{64.0}$^\ast$ & \textbf{12.4} \\
\midrule
\multicolumn{5}{l}{\emph{Lens change}} \\
frozen & 82.4 & 82.4 & 53.2 & 5.1 \\
+\,\method{} & \textbf{88.4}$^\ast$ & \textbf{88.1}$^\ast$ & \textbf{65.9}$^\ast$ & \textbf{34.0}$^\ast$ \\
\bottomrule
\end{tabular}

\caption{\textbf{Other camera changes.} Success rate (\%) pooled over suites: CamRand on all four (1,200 episodes), AnyCamVLA on Goal, Spatial and Object (900), drift and lens change on all four (1,200 and 800). $^\ast$\,McNemar $p<0.05$.}
\label{tab:bench}
\end{table}

\subsection{Generalization to Diverse Camera Shifts}
\label{sec:bench}
\paragraph{One pipeline for 6-DoF shifts, lens swaps, and drift.}
Under \emph{CamRand} (\cref{tab:bench}), \method{} lifts the first three VLAs by $+3.3$ to $+9.8$ points and more than doubles OpenVLA (30.5\% $\to$ 62.8\%).
On \emph{AnyCamVLA-style orbits} across Goal, Spatial, and Object, it yields $+8.9$ to $+20.4$ point gains.
Under \emph{Lens Change} ($45^\circ \to 35^\circ/55^\circ$ FoV at fixed pose), VGGT's per-frame intrinsics $K_t$ correct perspective scaling, adding $+5.7$ to $+12.7$ points on the stronger policies and $+28.9$ points on OpenVLA.
Under \emph{Camera Drift}, cached poses grow stale, whereas \emph{per-step} pose tracking follows the moving camera to recover $+13.7$ to $+15.0$ points on OpenVLA-OFT, \pifive{}, and \pizero{} (OpenVLA does not improve under drift).
Finally, pairing large random camera shifts with LIBERO-Plus's 2,400 non-camera perturbations (including object layout) yields $+11$ to $+21$ point improvements per policy, with gains in all six perturbation categories (supplement).

\begin{table}[t]
\centering\footnotesize\setlength{\tabcolsep}{3.1pt}
\begin{tabular}{lcccccc}
\toprule
& scene & moved & & & & \\
& training & camera & OFT & \pifive{} & \pizero{} & OpenVLA \\
\midrule
frozen policy & -- & -- & 62.5 & 76.5 & 68.0 & 3.6 \\
AnyCamVLA$^\ddagger$ & fine-tuned & true & \textbf{80.7} & 85.1 & 87.5 & 26.0 \\
AnyCamVLA$^\S$ & fine-tuned & estimated & 79.5 & 82.1 & 83.1 & 14.1 \\
\textbf{\method{}} & \textbf{none} & \textbf{estimated} & 77.6 & \textbf{95.1} & \textbf{91.2} & \textbf{26.4} \\
\bottomrule
\end{tabular}

\caption{\textbf{Comparison with calibrated view synthesis} on all 1,599 LIBERO-Plus camera variants (success rate, \%). AnyCamVLA runs with the true ($^\ddagger$) or VGGT-estimated ($^\S$) parameters of the moved camera. Best per column in bold; per-suite results are in the supplementary material.}
\label{tab:vs}
\end{table}

\begin{table}[t]
\centering\footnotesize\setlength{\tabcolsep}{3pt}
\begin{tabular}{lcccc}
\toprule
& \multicolumn{2}{c}{OFT} & \multicolumn{2}{c}{OFT+ (fine-tuned)} \\
\cmidrule(lr){2-3}\cmidrule(lr){4-5}
& frozen & \textbf{+\,ours} & frozen & \textbf{+\,ours} \\
\midrule
LIBERO-Plus camera & 62.5 & \textbf{77.6}$^\ast$ & 95.7 & 91.1$^\ast$ \\
AnyCamVLA, Long & 33.7 & \textbf{61.7}$^\ast$ & 68.2 & \textbf{83.3}$^\ast$ \\
AnyCamVLA-style, other suites & 62.0 & \textbf{70.9}$^\ast$ & 84.2 & \textbf{87.4}$^\ast$ \\
CamRand & 83.8 & \textbf{93.6}$^\ast$ & 91.2 & \textbf{95.4}$^\ast$ \\
Drift (per-step pose) & 56.0 & \textbf{74.7}$^\ast$ & 76.5 & \textbf{86.7}$^\ast$ \\
Lens change & 80.2 & \textbf{82.2} & 83.0 & \textbf{86.0} \\
Clean camera & 94.8 & 94.8 & 96.2 & 96.0 \\
\bottomrule
\end{tabular}

\caption{\textbf{Restoration on top of a fine-tuned policy.} Success rate (\%) pooled as in \cref{tab:main,tab:bench}, except Drift (Goal, Long) and Lens change (Goal, Spatial), the suites released for OpenVLA-OFT+; drift uses the per-step variant, the clean camera the gated default. $^\ast$\,McNemar $p<0.05$ against the same policy frozen.}
\label{tab:oftplus}
\end{table}

\begin{figure}[t]
\centering
\includegraphics[width=0.92\linewidth]{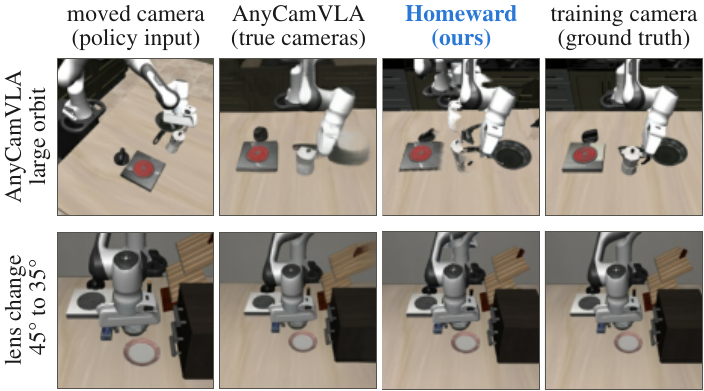}
\caption{\textbf{Qualitative canonicalization at matched simulator states} for a large AnyCamVLA orbit (Long) and an FoV narrowing ($45^\circ \to 35^\circ$). In frames from episodes where the frozen policy fails and \method{} succeeds, view synthesis blurs or omits fine objects, whereas \method{} preserves sharp scene geometry and crisp manipulator links.}
\label{fig:qual}
\end{figure}

\subsection{Comparison with Calibrated View Synthesis}
\label{sec:anycam}
We compare against AnyCamVLA~\cite{heo2026anycamvla}, which fine-tunes a view-synthesis network on LIBERO-Plus renders and requires the true deployment camera.

\paragraph{Competitive or superior on LIBERO-Plus.}
Across all 1,599 LIBERO-Plus variants (\cref{tab:vs}), training-free, uncalibrated \method{} \emph{outperforms} AnyCamVLA---even when AnyCamVLA is given \textbf{ground-truth} camera poses---by $+9.9$ points on \pifive{} ($[3.5, 16.8]$) and $+3.8$ points on \pizero{} ($[0.3, 7.5]$), matches it on OpenVLA ($+0.4$, $[-3.2, 4.1]$), and trails by only 3.1 points on OpenVLA-OFT ($[-11.3, 4.8]$); \cref{fig:qual} compares the restored views.
Under an equal uncalibrated setting where AnyCamVLA also consumes VGGT-estimated poses, \method{} leads by $+8.1$ to $+12.9$ points across \pifive{}, \pizero{}, and OpenVLA (and is within $-1.9$ points, $[-9.2, 5.3]$, on OpenVLA-OFT).

\paragraph{Characterizing the wide-orbit boundary.}
On AnyCamVLA's native wide-orbit benchmark on LIBERO-Long, \method{} improves frozen policies by $+28$ to $+33$ points, yet trails ground-truth-calibrated AnyCamVLA by 15.2, 4.8, and 5.8 points on OpenVLA-OFT, \pifive{}, and \pizero{}, with the gap widening at larger orbits.
Two controlled experiments isolate the mechanism (supplement): (i) without its fine-tuning on the benchmark scenes, AnyCamVLA's synthesis pipeline collapses \emph{below} the uncorrected policy; and (ii) supplying oracle depth and poses to \method{} closes the gap to within 3 points---or surpasses AnyCamVLA---even at the widest orbit.
Diagnostics confirm that while VGGT's median rotation error stays near $\sim\!1^\circ$, its median translation error grows from 2\,cm at small orbits to 7\,cm at large orbits, pointing to wide-baseline translation drift as the bottleneck.

\begin{table*}[t]
\centering\small\setlength{\tabcolsep}{5pt}
\begin{tabular}{lcccccccc}
\toprule
& \multicolumn{2}{c}{OFT} & \multicolumn{2}{c}{\pifive{}} & \multicolumn{2}{c}{\pizero{}} & \multicolumn{2}{c}{OpenVLA} \\
\cmidrule(lr){2-3}\cmidrule(lr){4-5}\cmidrule(lr){6-7}\cmidrule(lr){8-9}
& Goal & Long & Goal & Long & Goal & Long & Goal & Long \\
\midrule
frozen policy & 79.7 & 35.3 & 84.3 & 52.7 & 65.2 & 44.9 & 8.1 & 1.2 \\
\midrule
scene restoration only & 72.8$^\ast$ & 51.3$^\ast$ & 76.0$^\ast$ & 60.4$^\ast$ & 82.6$^\ast$ & 62.5$^\ast$ & 22.3$^\ast$ & 7.2$^\ast$ \\
\quad + whole robot incl.\ pedestal & 39.0$^\ast$ & 31.0 & 72.3$^\ast$ & 53.9 & 74.3$^\ast$ & 43.4 & 4.4$^\ast$ & 0.0 \\
\quad + moving links, pose per step & 88.2$^\ast$ & 70.6$^\ast$ & 93.6$^\ast$ & 83.1$^\ast$ & 90.7$^\ast$ & 74.0$^\ast$ & 36.3$^\ast$ & 12.2$^\ast$ \\
\quad + moving links, pose once & 90.0$^\ast$ & 71.4$^\ast$ & 93.4$^\ast$ & 92.6$^\ast$ & 92.9$^\ast$ & 73.5$^\ast$ & 35.8$^\ast$ & 11.0$^\ast$ \\
\quad \textbf{+ depth-edge filter (\method{})} & \textbf{91.4}$^\ast$ & \textbf{72.3}$^\ast$ & \textbf{94.4}$^\ast$ & \textbf{94.0}$^\ast$ & \textbf{93.4}$^\ast$ & \textbf{78.3}$^\ast$ & \textbf{39.2}$^\ast$ & \textbf{11.2}$^\ast$ \\
\quad\quad $\pi^3$ instead of VGGT & 88.0$^\ast$ & 74.5$^\ast$ & 94.6$^\ast$ & 89.5$^\ast$ & 94.9$^\ast$ & 74.5$^\ast$ & -- & -- \\
\midrule
reference frame + robot, no restoration & 70.8$^\ast$ & 41.5 & 78.2$^\ast$ & 63.0$^\ast$ & 85.5$^\ast$ & 69.2$^\ast$ & 11.3 & 0.0 \\
\method{} (per-step), no hole filling & 84.6 & 74.5$^\ast$ & 95.6$^\ast$ & 80.0$^\ast$ & 90.2$^\ast$ & 74.2$^\ast$ & 29.2$^\ast$ & 12.4$^\ast$ \\
\method{} (per-step), other reference frame & 89.0$^\ast$ & 72.6$^\ast$ & 94.1$^\ast$ & 82.3$^\ast$ & 93.1$^\ast$ & 75.9$^\ast$ & 36.3$^\ast$ & 13.1$^\ast$ \\
\method{} (per-step), reference of another task & 87.3$^\ast$ & 49.2$^\ast$ & 95.3$^\ast$ & 67.3$^\ast$ & 91.4$^\ast$ & 66.8$^\ast$ & 34.3$^\ast$ & 8.6$^\ast$ \\
\bottomrule
\end{tabular}

\caption{\textbf{Ablation study.} Success rate (\%) on the LIBERO-Plus camera variants of Goal (408) and Long (419). Rows marked (per-step) change one component of the per-step variant. --\,: not run. $^\ast$\,McNemar $p<0.05$ against the frozen policy.}
\label{tab:ablation}
\end{table*}

\subsection{Complementarity with Policy Fine-Tuning}
\label{sec:finetuned}
We next evaluate OpenVLA-OFT+~\cite{fei2025liberoplus}, obtained by fine-tuning OpenVLA-OFT on $>$20,000 perturbed LIBERO-Plus trajectories (\cref{tab:oftplus}).

\paragraph{Complementary gains on out-of-distribution shifts.}
On the LIBERO-Plus camera split---which lies close to OpenVLA-OFT+'s fine-tuning distribution (95.7\% base success)---test-time restoration is unnecessary and reduces success by 4.7 points ($[-10.1, -0.1]$, to 91.1\%).
However, across \textbf{all five out-of-distribution camera perturbation regimes} absent from fine-tuning, plugging in \method{} consistently boosts OpenVLA-OFT+ by $+3.0$ to $+15.1$ points---including $+15.1$ points on Long orbits ($[8.1, 23.7]$), $+10.2$ points under camera drift ($[2.7, 19.2]$), and $+4.2$ points on CamRand ($[1.2, 7.0]$), while leaving clean-camera performance intact ($-0.2$ points, $[-1.2, 0.5]$).

\subsection{Ablation Study}
\label{sec:mechanism}
\paragraph{Necessity of selective kinematic link rendering.}
Adding moving-link rendering and the 3\% depth-edge filter to \emph{scene restoration only} (\cref{tab:ablation}) yields dramatic jumps ($+18.4$ to $+33.6$ points for OpenVLA-OFT and \pifive{}; the filter alone contributes $+0.5$ to $+4.8$ points) and matches or exceeds \emph{oracle re-rendering} (\cref{tab:prelim}) in 6 of 8 suite--policy cells---because kinematic rendering eliminates arm disocclusion shadows that afflict even privileged point-cloud reprojection.

\paragraph{Impact of estimated \vs oracle scene geometry.}
Pairing our kinematic robot layer with simulator scene geometry raises OpenVLA-OFT across all 1,599 variants from 77.6\% to 90.2\% (driven largely by LIBERO-Object: 94.9\% \vs 50.3\%, a suite where the same estimated geometry already carries \pifive{} and \pizero{} to 93.9\% and 95.7\%), whereas \pifive{} and \pizero{} change by only 1.3 and 1.2 points overall (supplement).

\paragraph{Robustness to the geometry backbone.}
Replacing VGGT with $\pi^3$~\cite{wang2025pi3} alters performance across the six primary cells by only $-4.5$ to $+2.2$ points, achieving 76.7\% \vs 77.6\% overall on OpenVLA-OFT across all 1,599 variants ($p=0.41$). Sampled pixel MAE against ground-truth training frames is 4.5--6.0 intensity levels---matching oracle reprojection (4.6--6.2) and below uncorrected views (32--55).

\paragraph{Live reprojection, pose caching, and reference choice.}
Compositing the kinematic arm onto a static $I_{\mathrm{ref}}$ \emph{without} live scene reprojection underperforms \method{} in 14 of 16 cells (13 significantly; \eg, 41.5\% \vs 72.3\% for OFT on Long). Caching initial pose $(K_t,E_t)$ performs within 2.2 points of per-step estimation in 5 of 6 cells while surging $+9.5$ points ahead for \pifive{} on Long (92.6\% \vs 83.1\%), where per-step pose noise accumulates. Varying $I_{\mathrm{ref}}$ initial states or disabling hole filling shifts success by $\le 4$ points; using a cross-task reference from the same room costs $\le 0.9$ points on Goal, though 7--21 points on Long where scene object layouts change across tasks.

\subsection{Zero-Shift Safety and Seed Stability}
\label{sec:noharm}
\paragraph{Do-no-harm under unmoved cameras.}
On 400 clean-camera episodes per policy, the motion gate passes raw observations through in 99.25\% (1,985/2,000) of rollouts, keeping success within $\pm 0.5$ points ($[-2.2, 1.8]$).
Without the gate, restoring clean views alters success by at most 1.8 points ($[-6.8, 3.8]$), and on LIBERO-Plus's 2,400 non-camera variants by $-0.2$ to $+3.8$ points (all significant deltas positive).
Over three seeds on all 1,599 LIBERO-Plus variants, \pifive{} rises from $76.2 \pm 0.5\%$ to $94.6 \pm 0.5\%$ and \pizero{} from $67.7 \pm 1.6\%$ to $90.5 \pm 0.9\%$ (mean $\pm$ s.d.; supplement).

\section{Conclusion}
We investigated how far frozen VLA policies can be recovered from uncalibrated third-person camera shifts without any training.
Closed-loop diagnostics revealed that while viewpoint degradation is largely recoverable in observation space, feed-forward 3D models falter on thin, moving manipulator links---the foreground structure already known via robot kinematics.
By pairing monocular scene geometry with kinematic moving-link rendering, \method{} delivers consistent gains for OpenVLA-OFT, \pifive{}, and \pizero{} across five camera perturbation regimes and for OpenVLA in all but drift, matching or exceeding calibrated, scene-trained view synthesis on three of four policies on LIBERO-Plus.

\paragraph{Limitations and Future Work.}
Scene-trained synthesis retains an advantage under wide orbital baselines, where monocular translation estimation drifts, and restoration is unnecessary on viewpoints a fine-tuned policy has mastered.
Extending \method{} from simulation to physical robots is an important next step---requiring robustness to URDF-to-real photometric gaps, foreground occlusions in front of the manipulator, and contact-rich gripper--object interactions.
\FloatBarrier

\section*{LLM Usage Statement}
LLM-based tools were used extensively in this work, under the authors' direction and review.
\emph{Tools.} Claude Code (Anthropic) was the main coding and writing assistant; OpenAI Codex was used for review, auditing and manuscript preparation.
\emph{Research.} The tools searched and summarized related work; wrote and debugged most of the restoration, evaluation, experiment-orchestration and analysis code; launched and monitored the simulation runs; proposed controls, ablations and analyses, among which the authors chose; and audited the consistency of claims, statistics and tables.
\emph{Writing and figures.} The paper and its supplement were drafted and revised with these tools from the authors' instructions. Plots and tables were produced with AI-assisted scripts. Figures~1 and~2 were assembled in LaTeX/TikZ from recorded simulator images and MuJoCo-rendered robot illustrations; the camera icon in Figure~2 is from 3dicons.
\emph{Results.} All reported success rates and statistical comparisons are computed from logged closed-loop episodes; no simulator observation or trial outcome was generated by an LLM.
The authors directed the work, decided which experiments to run and which claims to make, and are responsible for all content.

\clearpage
\setcounter{figure}{0}
\setcounter{table}{0}
\renewcommand{\theHfigure}{supp.\arabic{figure}}
\renewcommand{\theHtable}{supp.\arabic{table}}
\renewcommand{\thefigure}{S\arabic{figure}}
\renewcommand{\thetable}{S\arabic{table}}

\raggedbottom
\twocolumn[{%
\renewcommand\twocolumn[1][]{#1}%
\begin{center}\Large\bfseries Supplementary Material\end{center}
\begin{center}
\captionsetup{hypcap=false}
\centering
\includegraphics[width=0.75\linewidth]{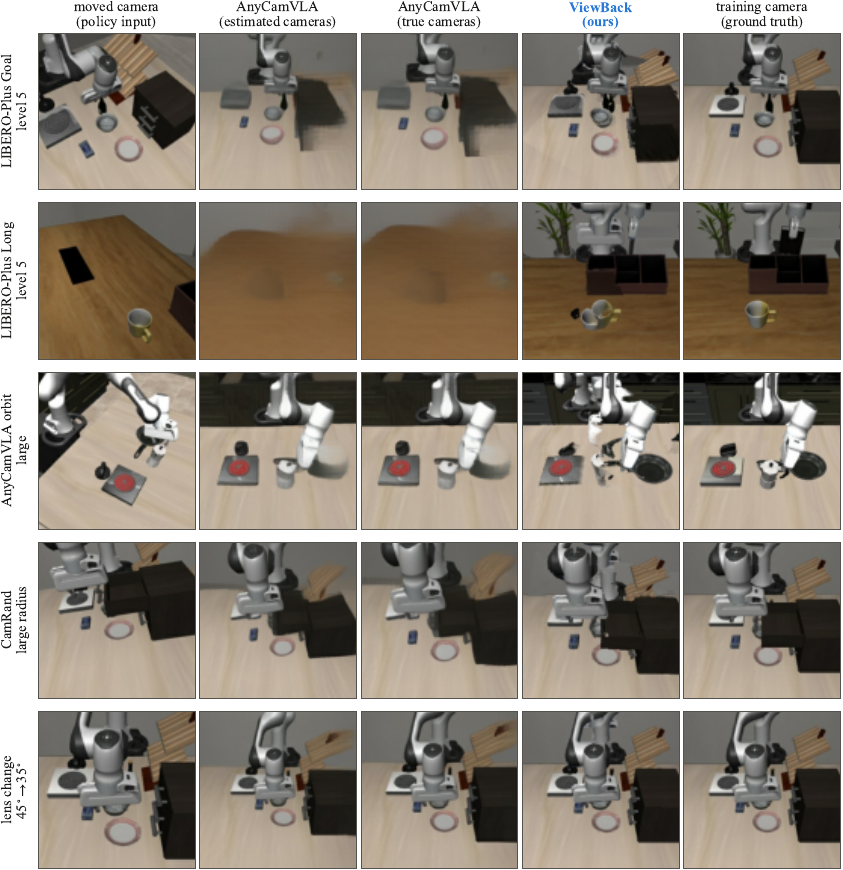}
\captionof{figure}{\textbf{Restored views of the same simulator state} for all five kinds of camera change, OpenVLA-OFT driven by \method{}.}
\label{fig:supp_grid}
\end{center}
}]
\appendix

\section{Qualitative results}
\label{sec:supp_qual}
\begin{figure*}[t]
\centering
\includegraphics[width=\linewidth]{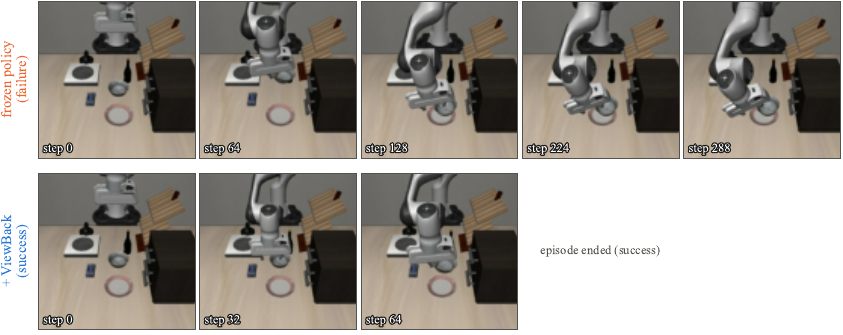}\\[0.4em]
\includegraphics[width=\linewidth]{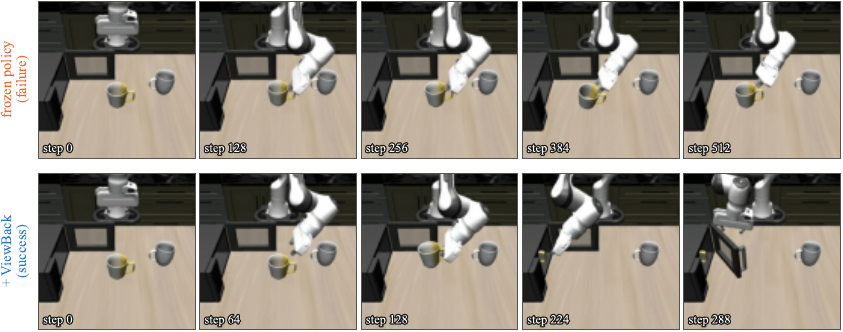}\\[0.4em]
\includegraphics[width=\linewidth]{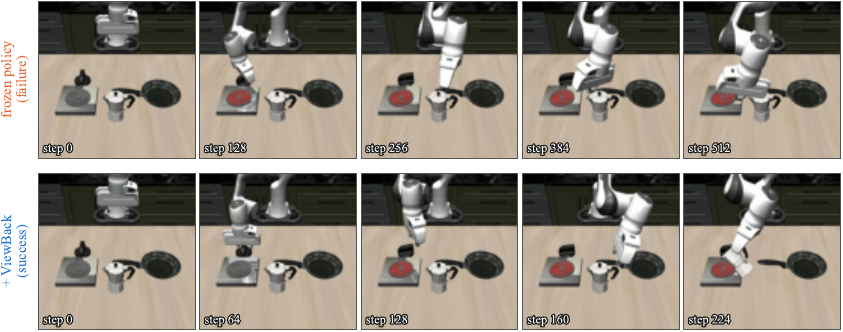}
\caption{\textbf{Episodes seen from the training camera}, frozen OpenVLA-OFT (top row of each pair) and with \method{} (bottom row), at evenly spaced queries of each episode; one query is eight control steps. Top: LIBERO-Plus Goal, level 5 (the scene of the main paper's first figure). Middle: LIBERO-Plus Long, level 5. Bottom: AnyCamVLA large orbit on Long.}
\label{fig:supp_rollouts}
\end{figure*}

\Cref{fig:supp_grid} shows restored views for all five kinds of camera change; view synthesis blurs or omits scene details, whereas \method{} preserves the observed structure and redraws the arm.
Every row shows one simulator state, mid-episode, as the moved camera sees it, as AnyCamVLA synthesizes the training view from it with estimated and with true camera parameters, as \method{} restores it, and as the training camera sees it.
\Cref{fig:supp_rollouts} follows three episodes through time from the training camera, without and with \method{}.
All episodes are drawn at random (fixed seed) among those where the frozen OpenVLA-OFT fails and \method{} succeeds, one per camera-shift level and benchmark; images are at the policy's input resolution.

\FloatBarrier
\twocolumn[{%
\begin{center}
\captionsetup{hypcap=false}
\centering\small\setlength{\tabcolsep}{6pt}
\begin{tabular}{lcccccc}
\toprule
Suite & frozen & ref.\ + robot & \textbf{\method{}} & AnyCam.$^\ddagger$ & AnyCam.$^\S$ & oracle + robot \\
\midrule
\multicolumn{7}{l}{\emph{OpenVLA-OFT}} \\
Spatial & 91.8 & 88.3 & \textbf{97.3}$^\ast$ & 98.1$^\ast$ & 99.5$^\ast$ & 91.5 \\
Object & 45.7 & 12.6$^\ast$ & \textbf{50.3} & 83.6$^\ast$ & 76.5$^\ast$ & 94.9$^\ast$ \\
Goal & 79.7 & 70.8$^\ast$ & \textbf{91.4}$^\ast$ & 81.4 & 85.0$^\ast$ & 91.4$^\ast$ \\
Long & 35.3 & 41.5 & \textbf{72.3}$^\ast$ & 61.6$^\ast$ & 58.9$^\ast$ & 83.3$^\ast$ \\
All & 62.5 & 52.8$^\ast$ & \textbf{77.6}$^\ast$ & 80.7$^\ast$ & 79.5$^\ast$ & 90.2$^\ast$ \\
\midrule
\multicolumn{7}{l}{\emph{\pifive{}}} \\
Spatial & 76.9 & 93.4$^\ast$ & \textbf{98.1}$^\ast$ & 78.2 & 74.2 & 98.1$^\ast$ \\
Object & 93.2 & 98.2$^\ast$ & \textbf{93.9} & 98.5$^\ast$ & 97.2$^\ast$ & 99.0$^\ast$ \\
Goal & 84.3 & 78.2$^\ast$ & \textbf{94.4}$^\ast$ & 91.7$^\ast$ & 89.2$^\ast$ & 97.1$^\ast$ \\
Long & 52.7 & 63.0$^\ast$ & \textbf{94.0}$^\ast$ & 72.3$^\ast$ & 68.0$^\ast$ & 91.9$^\ast$ \\
All & 76.5 & 82.7$^\ast$ & \textbf{95.1}$^\ast$ & 85.1$^\ast$ & 82.1$^\ast$ & 96.4$^\ast$ \\
\midrule
\multicolumn{7}{l}{\emph{\pizero{}}} \\
Spatial & 79.5 & 97.9$^\ast$ & \textbf{98.7}$^\ast$ & 95.2$^\ast$ & 93.6$^\ast$ & 97.9$^\ast$ \\
Object & 84.6 & 97.0$^\ast$ & \textbf{95.7}$^\ast$ & 97.2$^\ast$ & 94.9$^\ast$ & 96.5$^\ast$ \\
Goal & 65.2 & 85.5$^\ast$ & \textbf{93.4}$^\ast$ & 85.5$^\ast$ & 79.4$^\ast$ & 95.8$^\ast$ \\
Long & 44.9 & 69.2$^\ast$ & \textbf{78.3}$^\ast$ & 73.3$^\ast$ & 66.1$^\ast$ & 80.2$^\ast$ \\
All & 68.0 & 87.0$^\ast$ & \textbf{91.2}$^\ast$ & 87.5$^\ast$ & 83.1$^\ast$ & 92.4$^\ast$ \\
\midrule
\multicolumn{7}{l}{\emph{OpenVLA}} \\
Spatial & 3.2 & 22.6$^\ast$ & \textbf{32.2}$^\ast$ & 25.8$^\ast$ & 13.8$^\ast$ & 62.0$^\ast$ \\
Object & 2.0 & 0.0$^\ast$ & \textbf{23.7}$^\ast$ & 23.7$^\ast$ & 19.4$^\ast$ & 61.1$^\ast$ \\
Goal & 8.1 & 11.3 & \textbf{39.2}$^\ast$ & 39.2$^\ast$ & 19.4$^\ast$ & 52.7$^\ast$ \\
Long & 1.2 & 0.0 & \textbf{11.2}$^\ast$ & 15.5$^\ast$ & 4.1$^\ast$ & 25.1$^\ast$ \\
All & 3.6 & 8.2$^\ast$ & \textbf{26.4}$^\ast$ & 26.0$^\ast$ & 14.1$^\ast$ & 49.7$^\ast$ \\
\bottomrule
\end{tabular}

\captionof{table}{\textbf{Controls on all four suites.} Success rate (\%) on the LIBERO-Plus camera variants (376, 396, 408 and 419 per suite). Ref.\ + robot: the reference frame with the current robot drawn on it, no restoration. AnyCam.: AnyCamVLA's view synthesis with the true ($^\ddagger$) or VGGT-estimated ($^\S$) parameters of the moved camera, as in the main paper. Oracle + robot: re-rendering with the simulator's depth and true cameras plus the kinematic robot layer; not deployable. $^\ast$\,McNemar $p<0.05$ against the frozen policy.}
\label{tab:decomp}
\end{center}
}]
\section{Controls on all four suites}
\label{sec:supp_decomp}
The ablation table of the main paper covers Goal and Long; \cref{tab:decomp} runs its controls on all four LIBERO-Plus camera suites.
For OpenVLA, oracle re-rendering without the robot layer lifts Spatial from 3.2\% to 29.8\%, against 32.2\% for \method{}.
The reference frame with the current robot drawn on it is below \method{} in 14 of the 16 suite--policy cells, significantly in 13. It is above \method{} for \pifive{} on Object (98.2\% against 93.9\%, $p=0.003$) and, not significantly, for \pizero{} on Object (97.0\% against 95.7\%); on Object the oracle with the robot layer reaches 94.9--99.0\% for the first three policies, so the shortfall there lies in the estimated geometry.
Against AnyCamVLA with the true cameras, \method{} is ahead on Goal and Long for the first three policies (significantly in four of the six cells) and on Spatial for all but OpenVLA-OFT, where the two are on par (97.3\% against 98.1\%); it trails on Object for OpenVLA-OFT and \pifive{} and on Long for OpenVLA (11.2\% against 15.5\%, $p=0.025$).
With VGGT-estimated cameras, the information \method{} itself works from, AnyCamVLA is below \method{} in 13 of the 16 cells, significantly in 11; it is above on Object for OpenVLA-OFT and \pifive{} (76.5\% against 50.3\%, 97.2\% against 93.9\%) and on Spatial for OpenVLA-OFT (99.5\% against 97.3\%).

On AnyCamVLA's own protocol, run in our harness so both methods see the same episodes, the released model reaches 76.9\% with OpenVLA-OFT and 84.9\% with \pifive{}, close to the reported 78.9\% and 88.6\%~\cite{heo2026anycamvla}.
\Cref{tab:anycam} breaks LIBERO-Long down by orbit.
At the small orbit, \method{} is ahead of AnyCamVLA with true cameras for \pifive{} (92.0\% against 86.4\%) and \pizero{} (69.4\% against 68.4\%), and behind for OpenVLA-OFT (88.4\% against 89.6\%) and OpenVLA (8.0\% against 19.6\%); the calibrated model, fine-tuned on LIBERO-Plus renders, leads at the medium and large orbits.
Without that fine-tuning the same pipeline (zero-shot LVSM) falls below the frozen policy, whereas \method{} is trained on no scene.
A different 3D model does not change the picture: with $\pi^3$~\cite{wang2025pi3} the large orbit moves by at most a few points, and with the simulator's depth and cameras the same re-rendering reaches 70.0\%, 81.0\% and 63.6\% there, near or above AnyCamVLA, so the limit at wide orbits is the estimated geometry rather than content the moved camera never observed.

\begin{table}[!htbp]
\centering\small\setlength{\tabcolsep}{4pt}
\begin{tabular}{lcccc}
\toprule
LIBERO-Long & small & medium & large & all \\
\midrule
\multicolumn{5}{l}{\emph{OpenVLA-OFT}} \\
frozen & 80.6 & 9.6 & 11.0 & 33.7 \\
\textbf{+\,\method{}} & \textbf{88.4} & \textbf{62.6} & \textbf{34.0} & \textbf{61.7}$^\ast$ \\
+\,AnyCamVLA$^\ddagger$ & 89.6 & 79.4 & 61.6 & 76.9$^\ast$ \\
+\,AnyCamVLA$^\S$ & 85.6 & 78.8 & 57.8 & 74.1$^\ast$ \\
+\,zero-shot LVSM$^\ddagger$ & 40.8 & 18.2 & 2.6 & 20.5$^\ast$ \\
+\,zero-shot LVSM$^\S$ & 41.6 & 18.6 & 4.6 & 21.6$^\ast$ \\
\midrule
\multicolumn{5}{l}{\emph{\pifive{}}} \\
frozen & 86.4 & 55.6 & 9.0 & 50.3 \\
\textbf{+\,\method{}} & \textbf{92.0} & \textbf{79.0} & \textbf{69.2} & \textbf{80.1}$^\ast$ \\
+\,AnyCamVLA$^\ddagger$ & 86.4 & 84.6 & 83.6 & 84.9$^\ast$ \\
+\,AnyCamVLA$^\S$ & 86.8 & 83.2 & 80.8 & 83.6$^\ast$ \\
+\,zero-shot LVSM$^\ddagger$ & 70.0 & 17.8 & 14.8 & 34.2$^\ast$ \\
+\,zero-shot LVSM$^\S$ & 68.6 & 17.4 & 13.4 & 33.1$^\ast$ \\
\midrule
\multicolumn{5}{l}{\emph{\pizero{}}} \\
frozen & 66.0 & 16.6 & 5.4 & 29.3 \\
\textbf{+\,\method{}} & \textbf{69.4} & \textbf{65.4} & \textbf{51.4} & \textbf{62.1}$^\ast$ \\
+\,AnyCamVLA$^\ddagger$ & 68.4 & 69.6 & 65.6 & 67.9$^\ast$ \\
+\,AnyCamVLA$^\S$ & 64.4 & 66.8 & 65.6 & 65.6$^\ast$ \\
+\,zero-shot LVSM$^\ddagger$ & 49.6 & 10.8 & 7.0 & 22.5$^\ast$ \\
+\,zero-shot LVSM$^\S$ & 50.8 & 12.0 & 7.2 & 23.3$^\ast$ \\
\midrule
\multicolumn{5}{l}{\emph{OpenVLA}} \\
frozen & 7.6 & 0.0 & 0.0 & 2.5 \\
\textbf{+\,\method{}} & \textbf{8.0} & \textbf{2.6} & \textbf{0.0} & \textbf{3.5} \\
+\,AnyCamVLA$^\ddagger$ & 19.6 & 5.4 & 1.4 & 8.8$^\ast$ \\
+\,AnyCamVLA$^\S$ & 3.0 & 0.4 & 0.4 & 1.3$^\ast$ \\
\bottomrule
\end{tabular}

\caption{\textbf{AnyCamVLA protocol} on LIBERO-Long, 500 episodes per offset (10 tasks $\times$ 50 initial states). Our frozen OpenVLA-OFT and \pifive{} reach 33.7\% and 50.3\%, within 1.3 points of the 32.9\% and 49.0\% reported in~\cite{heo2026anycamvla}. $^\ddagger$\,AnyCamVLA's released view-synthesis model run in our harness with the true parameters of the moved third-person camera, the wrist camera and the training camera; it is fine-tuned on LIBERO-Plus renders. $^\S$\,The same model with the moved camera's pose and intrinsics estimated by VGGT against a training-view reference frame, as \method{} estimates them; the metric scale comes from the training view's depth, and the wrist and training cameras keep their true parameters. Zero-shot LVSM: the same pipeline with LVSM's public RealEstate10K weights~\cite{jin2024lvsm}, without AnyCamVLA's fine-tuning. \method{} requires no supplied calibration of the moved or wrist camera. $^\ast$\,McNemar $p<0.05$ against the frozen policy.}
\label{tab:anycam}
\end{table}

\section{Ghost arm and camera refinement}
\label{sec:supp_refine}
\method{} reprojects every pixel of the moved frame, the real arm included. When the estimated camera is off, this warped copy of the real arm appears beside the arm that the robot layer draws. On the AnyCamVLA protocol it pulls \pifive{} on Object at the medium orbit below the frozen policy (61.4\% against 76.0\%); with the simulator's depth and camera the same episodes reach 89.6\%.
We tested two fixes. Erasing the real arm before warping, with a mask of the moving links rendered from the estimated camera, lifts that cell to 88.4\% but costs \pifive{} 8.3 points on LIBERO-Plus Long, where the misplaced mask cuts objects next to the arm. Refining the camera first, by a cross-entropy search that matches the pixels the moving links change over up to seven frames of the episode, raises the median mask overlap on Goal and Long from 0.18 and 0.42 to 0.71 and 0.76.
With the refined camera the erasure keeps the gain for \pifive{} on Object but does not raise success in general (\cref{tab:refine}): it helps \pifive{} on the AnyCamVLA protocol, leaves \pizero{} unchanged and lowers OpenVLA on Object. \method{} therefore keeps the camera estimated from the first frame, and Object remains its weakest suite.

\begin{table}[!htbp]
\centering\small\setlength{\tabcolsep}{3pt}
\resizebox{\linewidth}{!}{\begin{tabular}{lcccc}
\toprule
& \multicolumn{2}{c}{All suites} & \multicolumn{2}{c}{Object} \\
\cmidrule(lr){2-3}\cmidrule(lr){4-5}
Policy & \method{} & + refined erase & \method{} & + refined erase \\
\midrule
\multicolumn{5}{l}{\emph{LIBERO-Plus camera variants}} \\
\pifive{} & 95.1 & 95.4 & 93.9 & 97.0$^\ast$ \\
\pizero{} & 91.2 & 90.3 & 95.7 & 97.0 \\
OpenVLA & 31.6$^\dagger$ & 27.7$^\ast$ & 23.7 & 12.1$^\ast$ \\
\midrule
\multicolumn{5}{l}{\emph{AnyCamVLA protocol, all orbits}} \\
\pifive{} & 78.5 & 82.7$^\ast$ & 76.8 & 89.3$^\ast$ \\
\pizero{} & 61.0 & 61.9 & 74.9 & 76.7 \\
OpenVLA & -- & -- & 8.5 & 0.3$^\ast$ \\
\bottomrule
\end{tabular}
}
\caption{\textbf{Erasing the real arm with a refined camera.} Success rate (\%) of \method{} without and with erasure of the real arm under a camera refined from the arm's motion, on the same episodes. All suites: the 1,599 LIBERO-Plus camera variants or the 6,000 AnyCamVLA episodes; $^\dagger$\,Goal and Object only (804). $^\ast$\,McNemar $p<0.05$ against \method{}.}
\label{tab:refine}
\end{table}

\section{Compound perturbations}
\label{sec:supp_compound}
LIBERO-Plus perturbs each suite along six factors other than the camera, with 100 variants each.
We combine each of these 2,400 variants with a CamRand camera at the largest radius, so that the policy faces a moved camera and a second perturbation at once.
\method{} raises all four policies in every category (\cref{tab:supp_compound}); it restores the camera, and the other factor stays as LIBERO-Plus sets it.

\begin{table*}[!t]
\centering\small\setlength{\tabcolsep}{3pt}
\resizebox{\linewidth}{!}{\begin{tabular}{lrcccccccc}
\toprule
& & \multicolumn{2}{c}{OFT} & \multicolumn{2}{c}{\pifive{}} & \multicolumn{2}{c}{\pizero{}} & \multicolumn{2}{c}{OpenVLA} \\
\cmidrule(lr){3-4}\cmidrule(lr){5-6}\cmidrule(lr){7-8}\cmidrule(lr){9-10}
Category & $n$ & frozen & \textbf{+\,\method{}} & frozen & \textbf{+\,\method{}} & frozen & \textbf{+\,\method{}} & frozen & \textbf{+\,\method{}} \\
\midrule
Background Textures & 400 & 62.2 & \textbf{80.5}$^\ast$ & 83.2 & \textbf{92.8}$^\ast$ & 69.0 & \textbf{84.8}$^\ast$ & 4.8 & \textbf{31.2}$^\ast$ \\
Language Instructions & 400 & 58.2 & \textbf{78.0}$^\ast$ & 75.2 & \textbf{88.5}$^\ast$ & 53.0 & \textbf{65.2}$^\ast$ & 3.0 & \textbf{23.8}$^\ast$ \\
Light Conditions & 400 & 69.8 & \textbf{89.0}$^\ast$ & 88.2 & \textbf{96.2}$^\ast$ & 75.5 & \textbf{89.8}$^\ast$ & 3.2 & \textbf{24.8}$^\ast$ \\
Objects Layout & 400 & 46.5 & \textbf{67.2}$^\ast$ & 76.5 & \textbf{84.5}$^\ast$ & 58.2 & \textbf{75.5}$^\ast$ & 4.5 & \textbf{34.2}$^\ast$ \\
Robot Initial States & 400 & 23.2 & \textbf{34.2}$^\ast$ & 66.5 & \textbf{80.5}$^\ast$ & 36.2 & \textbf{48.2}$^\ast$ & 2.5 & \textbf{11.8}$^\ast$ \\
Sensor Noise & 400 & 53.8 & \textbf{71.8}$^\ast$ & 79.5 & \textbf{93.5}$^\ast$ & 69.0 & \textbf{89.5}$^\ast$ & 0.8 & \textbf{19.0}$^\ast$ \\
\midrule
All & 2400 & 52.3 & \textbf{70.1}$^\ast$ & 78.2 & \textbf{89.3}$^\ast$ & 60.2 & \textbf{75.5}$^\ast$ & 3.1 & \textbf{24.1}$^\ast$ \\
\bottomrule
\end{tabular}
}
\caption{\textbf{Compound perturbations.} Success rate (\%) on the 2,400 non-camera variants of LIBERO-Plus (600 per suite), each under a CamRand camera at the largest radius, by perturbation category; one episode per variant. $^\ast$\,McNemar $p<0.05$ against the frozen policy.}
\label{tab:supp_compound}
\end{table*}

\section{Calibration error of the training camera}
\label{sec:supp_calib}
\method{} draws the moving links at the training camera, which it takes as calibrated once, before deployment.
Perturbing that calibration keeps every cell above its frozen baseline (\cref{tab:calib}); the margin is smallest for OpenVLA-OFT on Goal at 20\,mm and $2^\circ$ (86.0\% against 81.6\%, not significant).
Up to 10\,mm and $1^\circ$ of error no cell loses more than 3.2 points against the exact calibration. At 20\,mm and $2^\circ$ the largest losses are on Long: OpenVLA-OFT drops from 71.1\% to 58.9\% and \pifive{} from 92.4\% to 86.6\%, both still far above their frozen 34.8\% and 52.7\%.

\begin{table}[!htbp]
\centering\scriptsize\setlength{\tabcolsep}{2.7pt}
\begin{tabular}{lcccccc}
\toprule
& \multicolumn{2}{c}{OFT} & \multicolumn{2}{c}{\pifive{}} & \multicolumn{2}{c}{\pizero{}} \\
\cmidrule(lr){2-3}\cmidrule(lr){4-5}\cmidrule(lr){6-7}
& Goal & Long & Goal & Long & Goal & Long \\
\midrule
frozen policy & 81.6 & 34.8 & 83.8 & 52.7 & 65.2 & 44.9 \\
\method{}, exact training camera & 87.3$^\ast$ & 71.1$^\ast$ & 95.6$^\ast$ & 92.4$^\ast$ & 92.9$^\ast$ & 73.5$^\ast$ \\
\quad camera off by 5\,mm, $0.5^\circ$ & 89.0$^\ast$ & 69.5$^\ast$ & 94.9$^\ast$ & 89.7$^\ast$ & 92.9$^\ast$ & 74.2$^\ast$ \\
\quad camera off by 10\,mm, $1^\circ$ & 88.7$^\ast$ & 68.0$^\ast$ & 93.4$^\ast$ & 90.9$^\ast$ & 89.7$^\ast$ & 75.9$^\ast$ \\
\quad camera off by 20\,mm, $2^\circ$ & 86.0 & 58.9$^\ast$ & 93.4$^\ast$ & 86.6$^\ast$ & 90.7$^\ast$ & 75.2$^\ast$ \\
\bottomrule
\end{tabular}

\caption{\textbf{Calibration error of the training camera.} Success rate (\%) on the LIBERO-Plus camera variants of Goal (408) and Long (419) when the training camera that the robot is drawn from is off by the given translation and rotation (random direction and axis, fixed per episode). All rows run on one machine, whose renderer differs slightly from that of the ablation table of the main paper, and without the depth-edge filter. $^\ast$\,McNemar $p<0.05$ against the frozen policy.}
\label{tab:calib}
\end{table}

\section{Generic test-time fixes}
\label{sec:supp_generic}
\Cref{tab:generic} gives the numbers behind the comparison with temporal ensembling and 2D registration in the main paper; \cref{sec:supp_2dk} adds the robot layer to 2D registration.

\begin{table}[!htbp]
\centering\small\setlength{\tabcolsep}{4pt}
\begin{tabular}{lccc}
\toprule
& OFT & \pifive{} & \pizero{} \\
\midrule
frozen policy & 62.5 & 76.5 & 68.0 \\
+ temporal ensembling of action chunks & 64.7$^\ast$ & 74.8 & 67.3 \\
+ 2D registration to the reference frame & 62.4 & 74.1$^\ast$ & 70.7$^\ast$ \\
\textbf{+\,\method{}} & \textbf{77.6}$^\ast$ & \textbf{95.1}$^\ast$ & \textbf{91.2}$^\ast$ \\
\bottomrule
\end{tabular}

\caption{\textbf{Generic test-time fixes.} Success rate (\%) on all 1,599 LIBERO-Plus camera variants, pooled over the four suites. Temporal ensembling queries the policy every 4 control steps and averages the overlapping action chunks~\cite{zhao2023act}. 2D registration warps every frame onto the reference frame with a homography fitted once per episode to SIFT matches by RANSAC~\cite{lowe2004sift,fischler1981ransac}; like \method{}, it requires no supplied camera calibration. $^\ast$\,McNemar $p<0.05$ against the frozen policy.}
\label{tab:generic}
\end{table}

\section{2D alignment with the robot layer}
\label{sec:supp_2dk}
\Cref{tab:2dk} asks whether the 3D scene layer is needed once the robot is drawn from kinematics. It gives two 2D alignments the same kinematic robot layer as \method{}: the SIFT/RANSAC homography of \cref{tab:generic}, fitted once per episode, and the plane-induced homography computed from the true cameras.
The robot layer helps both alignments in 11 of 12 cells, by up to 27.2 points; 2D registration loses 6.9 points for OpenVLA-OFT on Goal.
\method{} stays ahead in all 12 comparisons, by 3.4 to 19.9 points on Goal and by 14.8 to 26.5 points on Long (McNemar $p\le 0.044$).

\begin{table}[!htbp]
\centering\small\setlength{\tabcolsep}{3.2pt}
\resizebox{\linewidth}{!}{\begin{tabular}{lcccccc}
\toprule
& \multicolumn{2}{c}{OFT} & \multicolumn{2}{c}{\pifive{}} & \multicolumn{2}{c}{\pizero{}} \\
\cmidrule(lr){2-3}\cmidrule(lr){4-5}\cmidrule(lr){6-7}
& Goal & Long & Goal & Long & Goal & Long \\
\midrule
frozen policy & 79.7 & 35.3 & 84.3 & 52.7 & 65.2 & 44.9 \\
\midrule
2D registration & 78.4 & 44.6$^\ast$ & 84.8 & 56.1 & 70.8$^\ast$ & 49.6$^\ast$ \\
\quad + robot layer & 71.6$^\ast$ & 51.3$^\ast$ & 89.7$^\ast$ & 67.5$^\ast$ & 87.5$^\ast$ & 61.8$^\ast$ \\
oracle homography$^\dagger$ & 70.8$^\ast$ & 38.7 & 74.8$^\ast$ & 51.3 & 60.8 & 41.5 \\
\quad + robot layer & 79.7 & 46.3$^\ast$ & 90.9$^\ast$ & 78.5$^\ast$ & 86.0$^\ast$ & 63.5$^\ast$ \\
\midrule
scene restoration only & 72.8$^\ast$ & 51.3$^\ast$ & 76.0$^\ast$ & 60.4$^\ast$ & 82.6$^\ast$ & 62.5$^\ast$ \\
\textbf{\method{}} & \textbf{91.4}$^\ast$ & \textbf{72.3}$^\ast$ & \textbf{94.4}$^\ast$ & \textbf{94.0}$^\ast$ & \textbf{93.4}$^\ast$ & \textbf{78.3}$^\ast$ \\
\bottomrule
\end{tabular}
}
\caption{\textbf{2D alignment with the kinematic robot layer.} Success rate (\%) on the LIBERO-Plus camera variants of Goal (408) and Long (419). $^\dagger$\,uses the true cameras; not deployable. $^\ast$\,McNemar $p<0.05$ against the frozen policy.}
\label{tab:2dk}
\end{table}

\section{Colour error of the robot layer}
\label{sec:supp_colour}
A mesh render of a real robot differs in colour from the camera image. We draw the robot layer with a per-episode gain and gamma error of size $S$ per channel, separately for light and dark parts, times a smooth shading field of amplitude $0.4S$; the colour-fit rows refit the colours once per episode on the reference frame. $S=50$ matches the error we measure for a real-robot render: about 30 grey levels in the robot region before the fit and 17 after it.
Success falls with $S$ for OpenVLA-OFT on Long (72.3\% to 57.3\% at $S=100$) and more slowly for \pifive{} (94.0\% to 88.3\%); \pizero{} on Long does not fall. The colour fit does not recover the loss consistently (\cref{tab:colour}).

\begin{table}[!htbp]
\centering\small\setlength{\tabcolsep}{3pt}
\resizebox{\linewidth}{!}{\begin{tabular}{lcccccccc}
\toprule
& \multicolumn{2}{c}{OFT} & \multicolumn{2}{c}{\pifive{}} & \multicolumn{2}{c}{\pizero{}$^\dagger$} & \multicolumn{2}{c}{OpenVLA} \\
\cmidrule(lr){2-3}\cmidrule(lr){4-5}\cmidrule(lr){6-7}\cmidrule(lr){8-9}
& Goal & Long & Goal & Long & Goal & Long & Spatial & Object \\
\midrule
frozen policy & 79.7 & 35.3 & 84.3 & 52.7 & 64.5 & 45.6 & 3.2 & 2.0 \\
exact render & 91.4$^\ast$ & 72.3$^\ast$ & 94.4$^\ast$ & 94.0$^\ast$ & 91.7$^\ast$ & 73.3$^\ast$ & 32.2$^\ast$ & 23.7$^\ast$ \\
$S=25$ & 91.9$^\ast$ & 67.1$^\ast$ & 92.6$^\ast$ & 91.4$^\ast$ & 94.6$^\ast$ & 78.5$^\ast$ & 31.6$^\ast$ & 20.7$^\ast$ \\
$S=50$ & 89.2$^\ast$ & 64.2$^\ast$ & 92.2$^\ast$ & 90.2$^\ast$ & 91.9$^\ast$ & 77.1$^\ast$ & 28.7$^\ast$ & 15.4$^\ast$ \\
$S=100$ & 87.5$^\ast$ & 57.3$^\ast$ & 90.7$^\ast$ & 88.3$^\ast$ & 90.7$^\ast$ & 81.4$^\ast$ & 28.7$^\ast$ & 13.4$^\ast$ \\
$S=50$, colour fit & 87.3$^\ast$ & 64.9$^\ast$ & 92.6$^\ast$ & 86.6$^\ast$ & 92.4$^\ast$ & 80.7$^\ast$ & 26.1$^\ast$ & -- \\
$S=100$, colour fit & 89.7$^\ast$ & 59.2$^\ast$ & 90.2$^\ast$ & 85.2$^\ast$ & 90.4$^\ast$ & 80.4$^\ast$ & 19.4$^\ast$ & 14.6$^\ast$ \\
\bottomrule
\end{tabular}
}
\caption{\textbf{Colour error of the robot layer.} Success rate (\%) on the LIBERO-Plus camera variants. $^\dagger$\,noise seed 9, paired with the frozen policy under the same seed. OpenVLA ran this study on Spatial and Object first. --\,: not run. $^\ast$\,McNemar $p<0.05$ against the frozen policy.}
\label{tab:colour}
\end{table}

On the Goal and Long camera variants used in the ablation table of the main paper, \cref{tab:colour_abl} lists the same control at $S=50$ for all four policies.
For the first three policies the colour error costs at most 2.2 points on Goal; on Long it costs OpenVLA-OFT 8.1 points and \pifive{} 3.8 and does not lower \pizero{}.

\begin{table}[!htbp]
\centering\small\setlength{\tabcolsep}{3pt}
\resizebox{\linewidth}{!}{%
\begin{tabular}{lcccccccc}
\toprule
& \multicolumn{2}{c}{OFT} & \multicolumn{2}{c}{\pifive{}} & \multicolumn{2}{c}{\pizero{}} & \multicolumn{2}{c}{OpenVLA} \\
\cmidrule(lr){2-3}\cmidrule(lr){4-5}\cmidrule(lr){6-7}\cmidrule(lr){8-9}
& Goal & Long & Goal & Long & Goal & Long & Goal & Long \\
\midrule
\method{}, exact render & 91.4$^\ast$ & 72.3$^\ast$ & 94.4$^\ast$ & 94.0$^\ast$ & 93.4$^\ast$ & 78.3$^\ast$ & 39.2$^\ast$ & 11.2$^\ast$ \\
robot drawn with a real-robot colour error & 89.2$^\ast$ & 64.2$^\ast$ & 92.2$^\ast$ & 90.2$^\ast$ & 91.7$^\ast$ & 79.5$^\ast$ & 30.4$^\ast$ & 10.0$^\ast$ \\
\quad + colour fit on the reference frame & 87.3$^\ast$ & 64.9$^\ast$ & 92.6$^\ast$ & 86.6$^\ast$ & 93.9$^\ast$ & 79.7$^\ast$ & 21.8$^\ast$ & 13.6$^\ast$ \\
\bottomrule
\end{tabular}}
\caption{\textbf{Colour error on the suites of the main ablation table.} Success rate (\%) on the LIBERO-Plus camera variants of Goal (408) and Long (419), in the layout of the ablation table of the main paper. $^\ast$\,McNemar $p<0.05$ against the frozen policy.}
\label{tab:colour_abl}
\end{table}

\section{Unchanged camera}
\label{sec:supp_noharm}
\Cref{tab:noharm} gives the results behind the unchanged-camera paragraph of the main paper: 400 clean-camera episodes per policy, with the motion gate (default) and with the gate disabled, and the 2,400 non-camera variants of LIBERO-Plus.

\begin{table}[!htbp]
\centering\small\setlength{\tabcolsep}{4pt}
\begin{tabular}{lcccc}
\toprule
& OFT & \pifive{} & \pizero{} & OpenVLA \\
\midrule
\multicolumn{5}{l}{\emph{clean camera, 400 episodes}} \\
frozen & 94.8 & 96.8 & 92.5 & 75.2 \\
\textbf{+\,\method{}} & 94.8 & 96.2 & \textbf{92.8} & 74.8 \\
\quad without motion gate & 95.8 & 97.0 & 92.8 & 73.5 \\
\midrule
\multicolumn{5}{l}{\emph{non-camera variants of the four suites, 2,400 episodes}} \\
frozen & 74.5 & 90.7 & 76.0 & 34.1 \\
\textbf{+\,\method{}} & \textbf{76.2}$^\ast$ & 90.5 & \textbf{77.5}$^\ast$ & \textbf{37.9}$^\ast$ \\
\quad without motion gate & 76.2$^\ast$ & 91.2 & 75.9 & 36.8$^\ast$ \\
\bottomrule
\end{tabular}

\caption{\textbf{Performance without a camera shift.} Success rate (\%) on 400 clean-camera episodes (100 per suite) and on the 2,400 LIBERO-Plus non-camera variants of the four suites (lighting, background, noise, layout, robot initial state, language; 600 per suite). $^\ast$\,McNemar $p<0.05$.}
\label{tab:noharm}
\end{table}

\section{Motion gate under drift}
\label{sec:supp_gate_drift}
In the cached variant the motion gate is decided once, at the first policy query, and kept for the rest of the episode. Every displaced-camera result of the main paper, drift included, is run with the gate disabled.
The drift camera starts at the training pose at reset and reaches its target after 120 control steps; the first query follows 10 settling steps, when the camera has covered 8\% of its path.
\Cref{tab:gate_drift} compares three settings of the per-step variant on drift.
\emph{First query}: from the first-query estimates logged by the ungated runs, a gate decided once (rotation and translation criteria) would fire in 95--96\% of the 400 small-drift episodes, whose target is a $10^\circ$ orbit, and in none of the 800 others; these columns replace the outcome of each such episode by that of the frozen policy and are computed, not run.
\emph{Every query}: a separate run in which the gate, with all three criteria, is decided again at each query. On small drifts it passes the first query through in 96\% of the episodes and 8--10\% of all queries, after which restoration takes over; on larger drifts it never fires.
Against the ungated runs it changes pooled drift success by $+0.5$ $[-0.6, 1.7]$, $-1.8$ $[-3.5, -0.2]$ and $-0.2$ $[-2.6, 1.8]$ points for OpenVLA-OFT, \pifive{} and \pizero{} (intervals resample tasks).
On the 400 clean-camera episodes the same per-query gate passes 92--93\% of the queries through and scores 94.8\%, 96.0\% and 92.0\%, against 94.8\%, 96.8\% and 92.5\% for the frozen policies; no difference is significant.

\begin{table}[!htbp]
\centering\small\setlength{\tabcolsep}{2.6pt}
\resizebox{\linewidth}{!}{\begin{tabular}{lcccccccc}
\toprule
& \multicolumn{4}{c}{small drift (400)} & \multicolumn{4}{c}{all drift (1,200)} \\
\cmidrule(lr){2-5}\cmidrule(lr){6-9}
& frozen & no gate & \shortstack{first\\query} & \shortstack{every\\query} & frozen & no gate & \shortstack{first\\query} & \shortstack{every\\query} \\
\midrule
OpenVLA-OFT & 85.2 & 85.2 & 84.8 & 86.8 & 58.1 & 72.5 & 72.3 & 73.0 \\
\pifive{} & 84.8 & 90.0 & 84.8 & 88.0 & 67.5 & 82.5 & 80.8 & 80.8 \\
\pizero{} & 64.8 & 66.8 & 64.2 & 69.0 & 50.3 & 64.0 & 63.2 & 63.8 \\
OpenVLA & 30.8 & 23.8 & 30.2 & -- & 11.4 & 12.4 & 14.6 & -- \\
\bottomrule
\end{tabular}
}
\caption{\textbf{Motion gate under drift.} Success rate (\%) of the per-step variant. First query: computed from the ungated and the frozen runs of the same episodes. Every query: run with the gate decided at each query; OpenVLA was not run. On medium and large drifts the first-query gate never fires.}
\label{tab:gate_drift}
\end{table}

\section{Computation delay}
\label{sec:supp_delay}
The simulation is synchronous: physics advances only when the environment is stepped, after the policy and the restoration have returned, so compute time delays no observation in the experiments of the main paper.
\Cref{tab:delay} adds a delay of 5 control steps, 0.25\,s at the 20\,Hz control rate and above the 0.21\,s that \method{} adds per query, in two ways.
\emph{Hold}: inference blocks, as in a controller that waits for its policy; after each query the robot holds its pose for 5 steps (zero pose command, gripper command unchanged) while objects and, under drift, the camera keep moving, and then executes the chunk. These steps are not counted against the episode's step limit.
\emph{Old observations}: the robot never waits; at each query the policy receives the third-person image, wrist image and joint state of 5 steps earlier, and \method{} draws the robot layer for that moment, with no compensation.
Holding changes \method{} by $-3.7$ to $+1.2$ points on Goal and Long; the only significant change is \pifive{} on Goal, and every cell stays significantly above the frozen policy. Under drift, where the camera moves on during the pause, it changes the per-step variant by $-3.3$, $-1.3$ and $+0.3$ points.
Acting on old observations fails for every policy, with or without \method{}: frozen policies lose 15 to 48 points and restored ones 38 to 70. A chunk planned from a state that the robot has already left repeats motion it has executed, so this regime needs latency compensation in the policy; \method{} does not provide it.

\begin{table}[!htbp]
\centering\small\setlength{\tabcolsep}{2.6pt}
\resizebox{\linewidth}{!}{\begin{tabular}{lccccccccc}
\toprule
& \multicolumn{2}{c}{OFT} & \multicolumn{2}{c}{\pifive{}} & \multicolumn{2}{c}{\pizero{}} & \multicolumn{3}{c}{drift (1,200)} \\
\cmidrule(lr){2-3}\cmidrule(lr){4-5}\cmidrule(lr){6-7}\cmidrule(lr){8-10}
& Goal & Long & Goal & Long & Goal & Long & OFT & \pifive{} & \pizero{} \\
\midrule
frozen policy & 79.7 & 35.3 & 84.3 & 52.7 & 65.2 & 44.9 & 58.1 & 67.5 & 50.3 \\
\quad acting on 0.25\,s-old observations & 31.9$^\ast$ & 10.7$^\ast$ & 55.1$^\ast$ & 34.6$^\ast$ & 50.2$^\ast$ & 13.4$^\ast$ & -- & -- & -- \\
\midrule
\method{} & 91.4 & 72.3 & 94.4 & 94.0 & 93.4 & 78.3 & 72.5 & 82.5 & 64.0 \\
\quad robot holds 0.25\,s per query & 92.4 & 73.5 & 90.7$^\ast$ & 92.1 & 92.9 & 77.3 & 69.2$^\ast$ & 81.2 & 64.3 \\
\quad acting on 0.25\,s-old observations & 21.3$^\ast$ & 24.3$^\ast$ & 56.9$^\ast$ & 37.9$^\ast$ & 47.5$^\ast$ & 26.5$^\ast$ & -- & -- & -- \\
\bottomrule
\end{tabular}
}
\caption{\textbf{A 0.25\,s delay per policy query.} Success rate (\%) on the LIBERO-Plus camera variants of Goal (408) and Long (419) and on the drift benchmark (per-step variant). --\,: not run. $^\ast$\,McNemar $p<0.05$ against the same arm without delay.}
\label{tab:delay}
\end{table}

\section{Camera estimate and the Object suite}
\label{sec:supp_camerr}
\Cref{tab:camerr} compares the camera that VGGT estimates at the first query, from the reference frame and the first moved-camera frame, with the simulator's camera on all 1,599 LIBERO-Plus camera variants. Positions are put in metres with the depth of the training view; \method{} itself never needs this scale.
The median error is about $1^\circ$ in rotation and 5\,cm in position in every suite.
The exception is a group of 53 Object variants in which the camera is only pulled back, to 148--200\% of its training distance: there the position error exceeds 20\,cm. The 51 to 53 variants with the same pull-back in each other suite have a median error of 1.6 to 7.2\,cm and none above 12.5\,cm.

\begin{table}[!htbp]
\centering\small\setlength{\tabcolsep}{4pt}
\begin{tabular}{lccccc}
\toprule
& variants & \shortstack{rotation\\(deg)} & \shortstack{position\\(cm)} & \shortstack{focal\\(\%)} & \shortstack{position\\$>$20\,cm} \\
\midrule
Spatial & 376 & 1.2 / 3.0 & 5.2 / 9.4 & 5.2 / 14.1 & 0 \\
Object & 396 & 0.9 / 1.7 & 5.8 / 64.4 & 3.5 / 7.9 & 53 \\
Goal & 408 & 1.2 / 2.3 & 5.9 / 8.8 & 6.6 / 12.8 & 0 \\
Long & 419 & 0.9 / 1.7 & 4.5 / 7.5 & 4.7 / 13.6 & 0 \\
\midrule
All & 1,599 & 1.0 / 2.3 & 5.4 / 9.1 & 4.5 / 12.5 & 53 \\
\bottomrule
\end{tabular}

\caption{\textbf{First-query camera estimate against the simulator's camera} on the LIBERO-Plus camera variants: median / 90th percentile of the rotation error, the position error and the relative focal-length error, and the number of variants with a position error above 20\,cm.}
\label{tab:camerr}
\end{table}

\Cref{tab:object_split} splits the Object suite by this error.
On the 53 pulled-back variants \method{} does not help OpenVLA-OFT (5.7\%), lowers \pifive{} from 100.0\% to 67.9\%, and still lifts \pizero{} from 39.6\% to 90.6\%.
On the other 343 variants the camera estimate is as good as elsewhere and \pifive{} and \pizero{} reach 98.0\% and 96.5\%, yet OpenVLA-OFT gains only 5.2 points, to 57.1\%, against 98.8\% with the simulator's depth and cameras under the same robot layer and the same reference fill.
The gap of OpenVLA-OFT on Object is therefore not a matter of larger camera errors or of the reference frame: both are the same as in the suites, and for the policies, where restoration works. It is specific to this checkpoint, which is far more sensitive to the residual error of the estimated geometry than the two flow-matching policies. It appears at small shifts as well: at LIBERO-Plus levels 1 and 2 restoration lowers OpenVLA-OFT on Object from 88.9\% and 84.5\% to 66.7\% and 77.5\%.

\begin{table}[!htbp]
\centering\small\setlength{\tabcolsep}{3pt}
\resizebox{\linewidth}{!}{\begin{tabular}{lcccccc}
\toprule
& \multicolumn{3}{c}{position error $\le$ 20\,cm (343)} & \multicolumn{3}{c}{position error $>$ 20\,cm (53)} \\
\cmidrule(lr){2-4}\cmidrule(lr){5-7}
& frozen & \method{} & \shortstack{oracle geometry\\+ robot layer$^\dagger$} & frozen & \method{} & \shortstack{oracle geometry\\+ robot layer$^\dagger$} \\
\midrule
OpenVLA-OFT & 51.9 & 57.1 & 98.8$^\ast$ & 5.7 & 5.7 & 69.8$^\ast$ \\
\pifive{} & 92.1 & 98.0$^\ast$ & 98.8$^\ast$ & 100.0 & 67.9$^\ast$ & 100.0 \\
\pizero{} & 91.5 & 96.5$^\ast$ & 96.5$^\ast$ & 39.6 & 90.6$^\ast$ & 96.2$^\ast$ \\
\bottomrule
\end{tabular}
}
\caption{\textbf{Object suite by camera-estimate error.} Success rate (\%) on the LIBERO-Plus camera variants of Object. $^\dagger$\,simulator depth and cameras; not deployable. $^\ast$\,McNemar $p<0.05$ against the frozen policy.}
\label{tab:object_split}
\end{table}

\section{Failure cases}
\label{sec:supp_fail}
\begin{figure*}[!t]
\centering
\includegraphics[width=0.78\textwidth]{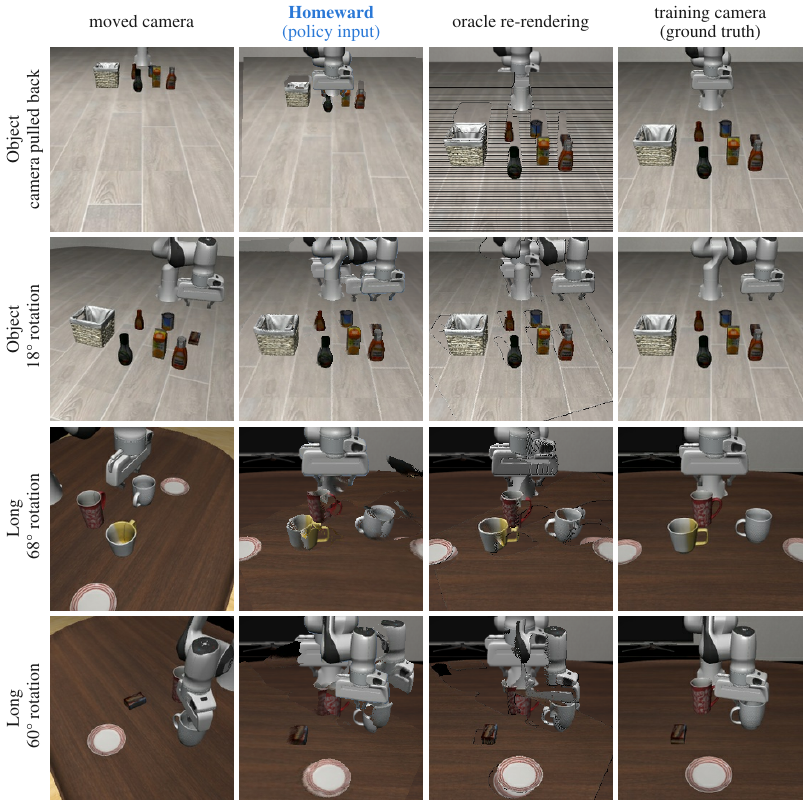}
\caption{\textbf{Failure cases of \method{}} with OpenVLA-OFT. Columns: the moved camera's frame, \method{}'s output, the oracle re-rendering from the simulator's depth and cameras (no robot layer), and the training camera's view of the same simulator state. Rows 1 and 3 show the first policy query, rows 2 and 4 the fifth and ninth. Rotations are those estimated at the first query.}
\label{fig:fail}
\end{figure*}

\Cref{fig:fail} shows four LIBERO-Plus episodes in which OpenVLA-OFT fails with \method{} and succeeds with the simulator's depth and cameras under the same robot layer.
In the first row the camera is pulled back to 182\% of its training distance, one of the 53 Object variants of \cref{sec:supp_camerr}: the estimate misses most of the pull-back (the camera position is off by 64\,cm), so the objects stay too small and too high while the robot layer is drawn at its true scale.
In the second row the camera estimate is accurate ($0.4^\circ$, 1.7\,cm), the restored view is as close to the training view as the oracle re-rendering (mean absolute error 7.0 against 7.4 grey levels) and the frozen policy succeeds on this variant, yet the policy fails with \method{}; what remains is a warped copy of the real arm beside the drawn one and softened small objects.
The third and fourth rows are rotations of $68^\circ$ and $60^\circ$ on Long: the moved camera sees the mugs from the other side, so their restored surfaces are torn and partly filled from the reference frame, and the warped real arm (\cref{sec:supp_refine}) covers the object in the gripper.

\section{Level 1 and wider motion gates}
\label{sec:supp_level1}
The 6.1-point loss of OpenVLA-OFT at Level 1 in the main paper falls on the Object suite. On its 63 Level-1 variants restoration lowers success from 88.9\% to 66.7\% (McNemar $p=0.007$); on the 136 Level-1 variants of the other three suites success goes from 95.6\% to 97.1\%. Pooled over the 199 variants the change is not significant ($p=0.058$). At the same level the other policies change by $+0.5$ (\pifive{}), $+6.5$ (\pizero{}, $p=0.007$) and $+36.6$ points (OpenVLA).

Level 1 is the mildest level of the benchmark, not an almost unmoved camera: by the first-query estimate its variants have a median rotation of $13.8^\circ$ and a median translation of 28\% of the scene depth (10th to 90th percentile: 0.9--$41.0^\circ$ and 1--56\%).
The motion gate tests for an unmoved camera and passes none of the 1,599 variants.
\Cref{tab:gate_sweep} emulates wider gates from the logs: an episode whose first-query estimate is below both the rotation and the translation threshold takes the outcome of the frozen run, every other episode the outcome of \method{}.
Up to four times the default thresholds no episode passes.
At ten times ($5^\circ$, 10\%) 35 episodes pass. On them restoration raises success from 42.9\% to 51.4\% for OpenVLA-OFT, from 60.0\% to 91.4\% for \pizero{} ($p=0.001$) and from 20.0\% to 25.7\% for OpenVLA, and lowers it from 94.3\% to 80.0\% for \pifive{} (2 episodes gained, 7 lost, $p=0.18$).
Wider gates pass a third to a half of the Level-1 variants. They leave OpenVLA-OFT at 85.9\% and 86.4\% there, and they lower pooled success significantly for every policy but \pifive{}, which stays within 0.3 points.
The observed Level-1 decline is concentrated on Object for this checkpoint (\cref{sec:supp_camerr}) and is not eliminated by the tested camera-motion gates.

\begin{table}[!htbp]
\centering\small\setlength{\tabcolsep}{3pt}
\resizebox{\linewidth}{!}{\begin{tabular}{lcccccccccc}
\toprule
& \multicolumn{2}{c}{passed (\%)} & \multicolumn{2}{c}{OFT} & \multicolumn{2}{c}{\pifive{}} & \multicolumn{2}{c}{\pizero{}} & \multicolumn{2}{c}{OpenVLA} \\
\cmidrule(lr){2-3}\cmidrule(lr){4-5}\cmidrule(lr){6-7}\cmidrule(lr){8-9}\cmidrule(lr){10-11}
gate thresholds & L1 & all & L1 & all & L1 & all & L1 & all & L1 & all \\
\midrule
no gate (\method{} as reported) & 0 & 0 & 87.4 & 77.6 & 98.0 & 95.1 & 97.0 & 91.2 & 47.2 & 26.4 \\
$0.5^\circ$, 1\% (default) & 0 & 0 & 87.4 & 77.6 & 98.0 & 95.1 & 97.0 & 91.2 & 47.2 & 26.4 \\
$2^\circ$, 4\% (4$\times$) & 0 & 0 & 87.4 & 77.6 & 98.0 & 95.1 & 97.0 & 91.2 & 47.2 & 26.4 \\
$5^\circ$, 10\% (10$\times$) & 3 & 2 & 86.9 & 77.4 & 98.0 & 95.4 & 96.5 & 90.6$^\ast$ & 46.7 & 26.3 \\
$10^\circ$, 20\% (20$\times$) & 31 & 14 & 85.9 & 76.0$^\ast$ & 98.0 & 95.4 & 96.0 & 88.6$^\ast$ & 27.6$^\ast$ & 19.8$^\ast$ \\
$15^\circ$, 30\% (30$\times$) & 45 & 24 & 86.4 & 74.2$^\ast$ & 98.0 & 95.0 & 93.5 & 86.0$^\ast$ & 22.1$^\ast$ & 14.8$^\ast$ \\
\midrule
frozen policy (all passed) & 100 & 100 & 93.5 & 62.5 & 97.5 & 76.5 & 90.5 & 68.0 & 10.6 & 3.6 \\
\bottomrule
\end{tabular}
}
\caption{\textbf{Wider motion gates, emulated} on the LIBERO-Plus camera variants. Success rate (\%) at Level 1 (L1, 199 variants) and over all 1,599 variants when episodes whose first-query estimate is below both thresholds (rotation; translation relative to the median scene depth) are passed through unrestored. Computed from the frozen and the \method{} runs of the same episodes; no threshold was tuned. Passed: share of episodes below the thresholds in the OpenVLA-OFT runs (the other runs differ by at most 0.5 points). $^\ast$\,McNemar $p<0.05$ against no gate.}
\label{tab:gate_sweep}
\end{table}

\section{Oracle re-rendering and the training camera}
\label{sec:supp_ceiling}
Oracle re-rendering is not the training camera's view. It reprojects the moved camera's frame with the simulator's depth and cameras and fills what that camera does not see from the reference frame, as \method{} does; only the geometry is privileged.
What the training camera itself renders of the same scene is the clean-camera condition, and \cref{tab:ceiling} sets the two side by side on Goal and Long.
With OpenVLA-OFT, \pifive{} and \pizero{}, the frozen policy at the training camera scores 5.3 to 14.9 points above oracle re-rendering (41.0 and 46.5 points with OpenVLA), so exact geometry alone does not return the training view.
The robot layer over the same oracle geometry comes within 7.7 points of the training camera for these three policies.
\method{}, with estimated geometry, is within 6.6 points in five of the six cells and 18.7 points below the training camera for OpenVLA-OFT on Long; OpenVLA stays 34.8 and 44.8 points below.
The clean-camera episodes use ten initial states per task and the camera variants one, so the first row is not paired with the others.

\begin{table}[!htbp]
\centering\small\setlength{\tabcolsep}{3pt}
\resizebox{\linewidth}{!}{\begin{tabular}{lcccccccc}
\toprule
& \multicolumn{2}{c}{OFT} & \multicolumn{2}{c}{\pifive{}} & \multicolumn{2}{c}{\pizero{}} & \multicolumn{2}{c}{OpenVLA} \\
\cmidrule(lr){2-3}\cmidrule(lr){4-5}\cmidrule(lr){6-7}\cmidrule(lr){8-9}
& Goal & Long & Goal & Long & Goal & Long & Goal & Long \\
\midrule
training camera, frozen policy$^\dagger$ & 98.0 & 91.0 & 96.0 & 91.0 & 95.0 & 84.0 & 74.0 & 56.0 \\
\midrule
moved camera, frozen policy & 79.7 & 35.3 & 84.3 & 52.7 & 65.2 & 44.9 & 8.1 & 1.2 \\
oracle re-rendering$^\ddagger$ & 90.4 & 76.1 & 88.7 & 77.8 & 89.7 & 74.7 & 27.5 & 15.0 \\
oracle re-rendering + robot layer$^\ddagger$ & 91.4 & 83.3 & 97.1 & 91.9 & 95.8 & 80.2 & 52.7 & 25.1 \\
\textbf{\method{}} & 91.4 & 72.3 & 94.4 & 94.0 & 93.4 & 78.3 & 39.2 & 11.2 \\
\bottomrule
\end{tabular}
}
\caption{\textbf{Training camera and oracle re-rendering.} Success rate (\%). $^\dagger$\,Clean-camera episodes: the unperturbed scene seen by the training camera, 100 per suite (10 tasks $\times$ 10 initial states). Other rows: the LIBERO-Plus camera variants of Goal (408) and Long (419), as in the ablation table of the main paper. $^\ddagger$\,Privileged: simulator depth and cameras.}
\label{tab:ceiling}
\end{table}

\section{Inference cost}
\label{sec:supp_cost}
\Cref{tab:cost} lists what the two view-restoration methods add to a policy query on one A100. Times are medians over the 1,599 \pifive{} episodes of the LIBERO-Plus camera variants (827 for the per-step variant); memory is the peak that the restoration models allocate in the evaluator process, measured in separate three-episode runs with the policy served from another process.
In the default variant the first query of an episode runs the two-frame estimate, the computation of a per-step query; later queries run VGGT on the current frame alone.
At inference AnyCamVLA is the cheaper method: its view synthesis is about five times faster than \method{}'s scene layer and, with the true cameras, takes a sixth of the memory. What it needs instead is a synthesis model fine-tuned on renders of the benchmark scenes and the parameters of the moved camera; when it estimates that camera with VGGT, as \method{} does, its memory is the larger of the two.

\begin{table}[!htbp]
\centering\small\setlength{\tabcolsep}{3pt}
\resizebox{\linewidth}{!}{%
\begin{tabular}{lccc}
\toprule
& models & \shortstack{time per\\query (ms)} & \shortstack{peak GPU\\memory (GiB)} \\
\midrule
\method{} & VGGT-1B & 164 + 10 & 6.6 \\
\method{}, per-step & VGGT-1B & 191 + 10 & 6.6 \\
AnyCamVLA, true cameras$^\ddagger$ & LVSM & 32 & 1.1 \\
AnyCamVLA, estimated cameras$^\S$ & LVSM, VGGT-1B & 31 & 7.2 \\
\bottomrule
\end{tabular}}
\caption{\textbf{Inference cost} of view restoration, excluding the policy. \method{}: scene layer + robot layer. $^\S$\,VGGT runs once per episode, on the first frame; its time is not included.}
\label{tab:cost}
\end{table}

\section{Prompt convention}
\label{sec:supp_prompt}
The LIBERO-Plus camera variants, CamRand and clean-camera episodes use base-task instructions; the other camera protocols use paraphrased instructions, as detailed in \cref{sec:supp_protocol}.
The client released with LIBERO-Plus builds the prompt from the variant's file name, which appends text such as ``view 0 0 100 2 352 initstate 0''; with that prompt our frozen OpenVLA-OFT, \pifive{} and \pizero{} score 58.2\%, 69.4\% and 14.8\% on the camera variants instead of 62.5\%, 76.5\% and 68.0\%, reproducing the 56.4\% and 13.8\% reported for OpenVLA-OFT and \pizero{}~\cite{fei2025liberoplus}.

\section{Renderer}
\begin{table}[t]
\centering\small\setlength{\tabcolsep}{3.5pt}
\begin{tabular}{lcccccc}
\toprule
& \multicolumn{2}{c}{frozen} & \multicolumn{2}{c}{\method{}} & \multicolumn{2}{c}{per-step} \\
\cmidrule(lr){2-3}\cmidrule(lr){4-5}\cmidrule(lr){6-7}
\pizero{} & GPU & software & GPU & software & GPU & software \\
\midrule
Spatial & 79.5 & 79.5 & 97.6 & 97.3 & 97.9 & 97.1 \\
Object & 84.6 & 85.1 & 96.0 & 96.2 & 96.7 & 96.2 \\
Goal & 65.2 & 64.7 & 92.9 & 93.1 & 90.7 & 91.9 \\
Long & 44.9 & 48.0 & 73.5 & 78.0$^\ast$ & 74.0 & 76.1 \\
\bottomrule
\end{tabular}

\caption{\textbf{Software rendering.} Success rate (\%) of \pizero{} on the LIBERO-Plus camera variants with MuJoCo rendered on the GPU (NVIDIA EGL, used for every other result) and in software (Mesa llvmpipe), on the same episodes. $^\ast$\,the two renderers differ, McNemar $p<0.05$.}
\label{tab:renderer}
\end{table}

\subsection{Software rendering}
\label{sec:renderer}
Simulated images depend slightly on the rasterizer.
Rendering the same \pizero{} episodes in software instead of on the GPU changes success by up to 4.5 points on Long (\cref{tab:renderer}), but the gain of \method{} is the same under both renderers: 27.7 and 28.4 points on Goal, 28.6 and 30.0 on Long.
Over all 37 \pizero{} arms we ran under both renderers, one differs at $p<0.05$, about what chance predicts. We nevertheless compare arms only within one renderer and report GPU rendering throughout.

\section{CamRand by radius}
\label{sec:supp_camrand}
On LIBERO-Long, \method{} raises OpenVLA-OFT from 87\% to 92\%, from 72\% to 91\% and from 43\% to 82\% at the small, medium and large CamRand radius (100 episodes each).

\section{Seeds}
\label{sec:supp_seeds}
On all 1,599 LIBERO-Plus camera variants, \pifive{} rises from $76.2\pm0.5\%$ to $94.6\pm0.5\%$ and \pizero{} from $67.7\pm1.6\%$ to $90.5\pm0.9\%$ over three seeds (mean $\pm$ s.d.; $p<10^{-50}$ for each seed).
Over three seeds on AnyCamVLA's orbits of all four suites, \pifive{} rises from $57.6\pm0.1\%$ to $78.8\pm0.3\%$ and \pizero{} from $39.6\pm0.2\%$ to $61.2\pm0.2\%$ with \method{} (mean $\pm$ s.d.; $p<10^{-50}$ for each seed), compared with $80.8\pm0.5\%$ and $60.0\pm0.5\%$ for AnyCamVLA with estimated cameras.

\section{Task-level resampling}
\label{sec:supp_cluster}
The camera variants of one LIBERO task share its scene and instruction, so the episodes of a task are not independent.
\Cref{tab:cluster} therefore resamples tasks instead of episodes on the main result: each bootstrap draw samples the 40 tasks (10 per suite) with replacement and keeps every variant of a sampled task (6 to 50 per task); intervals are the 2.5th and 97.5th percentiles of 20,000 draws.
Drawing the tasks within each suite gives slightly narrower intervals.
The sign test compares, over tasks, how many gain and how many lose episodes (ties dropped, two-sided); the Wilcoxon signed-rank test is taken over the 40 per-task differences in success rate (two-sided).

\begin{table}[!htbp]
\centering\scriptsize\setlength{\tabcolsep}{1.5pt}
\resizebox{\linewidth}{!}{%
\begin{tabular}{@{}lcccccc@{}}
\toprule
Policy & \shortstack{gain\\(points)} & \shortstack{95\% CI,\\tasks} & \shortstack{95\% CI, tasks\\within suite} & \shortstack{tasks\\up/down/tied} & \shortstack{sign\\test $p$} & \shortstack{Wilcoxon\\$p$} \\
\midrule
OpenVLA-OFT & $+15.1$ & $[7.6, 22.9]$ & $[9.0, 21.7]$ & 26 / 5 / 9 & $1.9\times10^{-4}$ & $4.5\times10^{-4}$ \\
\pifive{} & $+18.6$ & $[10.3, 27.1]$ & $[11.7, 25.6]$ & 23 / 9 / 8 & $0.020$ & $0.0053$ \\
\pizero{} & $+23.2$ & $[17.1, 29.7]$ & $[17.5, 28.9]$ & 35 / 1 / 4 & $1.1\times10^{-9}$ & $3.2\times10^{-7}$ \\
OpenVLA & $+22.8$ & $[17.7, 28.4]$ & $[18.4, 27.7]$ & 38 / 1 / 1 & $1.5\times10^{-10}$ & $8.4\times10^{-8}$ \\
\bottomrule
\end{tabular}}
\caption{\textbf{Task-level resampling of the main result.} Gain of \method{} over the frozen policy on all 1,599 LIBERO-Plus camera variants, with bootstrap intervals that resample the 40 tasks, and the number of tasks whose success count rises, falls or is unchanged.}
\label{tab:cluster}
\end{table}

The same resampling applies to the other comparisons of the main paper (\cref{tab:cluster2}); for AnyCamVLA's orbit protocol on LIBERO-Long there are 10 tasks.

\begin{table}[!htbp]
\centering\scriptsize\setlength{\tabcolsep}{2.5pt}
\begin{tabular}{@{}lrc@{}}
\toprule
Comparison (difference in points) & diff. & 95\% CI, tasks \\
\midrule
\multicolumn{3}{@{}l}{\emph{\method{} minus AnyCamVLA, LIBERO-Plus camera variants}} \\
OpenVLA-OFT & $-3.1$ / $-1.9$ & $[-11.3, 4.8]$ / $[-9.2, 5.3]$ \\
\pifive{} & $+9.9$ / $+12.9$ & $[3.5, 16.8]$ / $[6.4, 19.7]$ \\
\pizero{} & $+3.8$ / $+8.1$ & $[0.3, 7.5]$ / $[4.7, 12.5]$ \\
OpenVLA & $+0.4$ / $+12.3$ & $[-3.2, 4.1]$ / $[8.1, 16.7]$ \\
\midrule
\multicolumn{3}{@{}l}{\emph{\method{} minus AnyCamVLA (true cameras), orbit protocol on Long}} \\
OpenVLA-OFT & $-15.2$ & $[-22.7, -7.6]$ \\
\pifive{} & $-4.8$ & $[-10.9, 1.9]$ \\
\pizero{} & $-5.8$ & $[-10.3, -1.1]$ \\
OpenVLA & $-5.3$ & $[-8.3, -2.4]$ \\
\midrule
\multicolumn{3}{@{}l}{\emph{\method{} minus frozen policy, orbit protocol on Long}} \\
OpenVLA-OFT / \pifive{} & $+27.9$ / $+29.7$ & $[18.4, 37.0]$ / $[19.8, 39.0]$ \\
\pizero{} / OpenVLA & $+32.7$ / $+1.0$ & $[23.5, 42.2]$ / $[-2.4, 5.4]$ \\
\midrule
\multicolumn{3}{@{}l}{\emph{OpenVLA-OFT+ with \method{} minus OpenVLA-OFT+}} \\
LIBERO-Plus camera & $-4.7$ & $[-10.1, -0.1]$ \\
AnyCamVLA, Long & $+15.1$ & $[8.1, 23.7]$ \\
AnyCamVLA-style, other suites & $+3.2$ & $[-1.3, 8.3]$ \\
CamRand & $+4.2$ & $[1.2, 7.0]$ \\
Drift (per-step pose) & $+10.2$ & $[2.7, 19.2]$ \\
Lens change & $+3.0$ & $[-0.8, 7.0]$ \\
Clean camera & $-0.2$ & $[-1.2, 0.5]$ \\
\midrule
\multicolumn{3}{@{}l}{\emph{Clean camera, with the motion gate / gate disabled}} \\
OpenVLA-OFT & $0.0$ / $+1.0$ & $[-0.8, 0.8]$ / $[-1.2, 3.8]$ \\
\pifive{} & $-0.5$ / $+0.2$ & $[-1.5, 0.5]$ / $[-1.8, 2.5]$ \\
\pizero{} & $+0.2$ / $+0.2$ & $[-1.2, 1.8]$ / $[-2.5, 3.2]$ \\
OpenVLA & $-0.5$ / $-1.8$ & $[-2.2, 1.0]$ / $[-6.8, 3.0]$ \\
\bottomrule
\end{tabular}
\caption{\textbf{Task-level intervals for the other comparisons.} Differences in success rate (points, from unrounded rates) with 95\% bootstrap intervals that resample tasks (20,000 draws). Paired entries list true / estimated camera parameters.}
\label{tab:cluster2}
\end{table}

\section{Protocol notes}
\label{sec:supp_protocol}
The main evaluations use NVIDIA EGL; the software-rendering control in \cref{sec:renderer} additionally uses Mesa llvmpipe.
Reported results combine automatically regenerated tables and separately maintained controls. Methods were run on the same machine where possible, but 30 evaluation groups include runs from multiple machines. Task resampling does not account for systematic differences between machines.
The camera-change suites of OpenVLA exist as two complete runs on two machines, which agree on 85--100\% of the episodes and differ by at most 0.5 points in pooled success; we report the run of the machine that holds every arm of the camera suites.
The LIBERO-Plus camera variants, CamRand and the clean-camera episodes are prompted with the instruction of the base task.
The AnyCamVLA-protocol, drift and lens-change episodes are built on LIBERO-Plus files with paraphrased instructions and use those instructions; within each benchmark, all compared methods receive identical prompts.
Reference frames are rendered once per task from initial state 1, after 10 settling steps (from the next stored state when the episode itself starts from state 1). The LIBERO-Plus camera variants start from state 0 and the other benchmarks from states 0, 5, \dots, 45 or from all 50, so the objects and the arm in the reference frame are placed differently from the test episode.
The simulation is synchronous, so the compute time below delays no observation; \cref{sec:supp_delay} measures a delay of this size.
Scale alignment in the cached variant: the static pixels are found at VGGT's input resolution against the first moved-camera frame, and the scale stays at 1 when no more than 100 static pixels remain.
Timing (A100, VGGT-1B at 518\,px, bfloat16, medians over the LIBERO-Plus camera episodes): the scene layer alone, that is VGGT inference, depth-edge filtering, reprojection and reference fill, takes 0.15\,s per query with cached camera geometry and 0.19\,s when the camera is re-estimated at every query; the robot layer takes 0.01\,s.
The 0.21\,s in the main paper is an indirect estimate from the restored-minus-frozen wall-clock time per policy query under matched episode conditions (0.20 to 0.22\,s for the three policies that act in chunks). It also includes evaluation-harness overhead that is not timed separately; different rollout lengths introduce noise, so it is not a direct measurement of restoration latency. The 27\,ms amortization applies only to OpenVLA-OFT, \pifive{} and \pizero{}, which query once every eight steps.

\FloatBarrier

{\small
\bibliographystyle{ieeenat_fullname}
\bibliography{refs}
}
\end{document}